\documentclass{article} % For LaTeX2e
\usepackage{iclr2026_conference,times}

\usepackage{amsmath,amsfonts,bm}

\def\eqref#1{equation~\ref{#1}}
\def\1{\bm{1}}

\DeclareMathAlphabet{\mathsfit}{\encodingdefault}{\sfdefault}{m}{sl}
\SetMathAlphabet{\mathsfit}{bold}{\encodingdefault}{\sfdefault}{bx}{n}

\usepackage{hyperref}
\usepackage{url}
\usepackage{graphicx}
\usepackage{subcaption}
\usepackage{wrapfig}
\usepackage{booktabs}
\usepackage{pifont}
\usepackage[table]{xcolor}
\definecolor{green}{RGB}{0, 150, 0}
\definecolor{red}{RGB}{200, 0, 0}
\definecolor{orange_fig}{RGB}{197, 90, 17}
\definecolor{green_fig}{RGB}{84, 130, 53}
\usepackage{multirow}

\title{OmniVideo-100K: A Dataset for Audio-Visual Reasoning \\ through Structured Scripts and Evidence Chains}

\author{
Xinyue Cai$^{1}$,~Chaoyou Fu$^{1}$\thanks{Corresponding author.}~,~Yi-Fan Zhang$^{2}$,~Ran He$^{2}$,~Caifeng Shan$^{1}$ \\
$^{1}$Nanjing University \quad $^{2}$CASIA \\
\texttt{yzlmhzz@smail.nju.edu.cn}, \texttt{cyfu@nju.edu.cn}
}

\iclrfinalcopy % Uncomment for camera-ready version, but NOT for submission.
\begin{document}

\maketitle
% \thispagestyle{empty}
% \fancyhead[L]{}

\begin{abstract}
Current automated pipelines for audio-visual Question Answering (QA) generally adopt a ``video-caption-QA'' paradigm. However, these methods typically segment videos into short clips and generate separate descriptions for audio and visual modalities.
This decoupled processing severs inherent associations between sounds and their visual sources, while independent clip processing often causes inconsistent descriptions of the same entity across segments. Furthermore, coupling long-text comprehension and QA synthesis into a single step often restricts models to localized events, yielding questions lacking long-term temporal connections and deep cross-modal reasoning. To address these issues, we propose an automated data engine featuring two mechanisms:
(1) \textbf{Entity-Anchored Video Scripting} transforms videos into structured scripts, comprising summaries, main entity lists, and segment-wise audio-visual descriptions. The entity list serves as a global prior to ensure cross-segment referential consistency and reconstruct audio-visual associations.
(2) \textbf{Clue-Guided QA Generation} prompts models to first mine cross-segment, multimodal clues from the script, and subsequently generate QA pairs based on these high-value clues.
Leveraging this pipeline, we construct the instruction-tuning dataset \textbf{OmniVideo-100K} and a human-verified test set, \textbf{OmniVideo-Test}. 
Fine-tuning VITA-1.5, Qwen2.5-Omni-7B and Qwen3-Omni-30B on OmniVideo-100K yields performance gains of up to 20.59\% on OmniVideo-Test, demonstrating strong generalization (up to 12.64\% improvements) across established benchmarks like Daily-Omni and JointAVBench.
Code and data are available at https://github.com/MiG-NJU/OmniVideo-100K.
\end{abstract}

\section{Introduction}
In recent years, Multimodal Large Language Models (MLLMs) have evolved from vision-language to omni-modal.
Specifically, in the domain of video understanding, integrating the audio modality enables the joint analysis of audio-visual cues, thereby facilitating a deeper comprehension of video content~\citep{baichuan-omni-1.5,qwen3-omni,video-salmonn2,cat+,uniav}.
This advanced comprehension goes beyond merely processing visual and audio modalities in isolation;
% necessitating the effective modeling of complex cross-modal synergy.
instead, it requires effectively modeling complex cross-modal interactions.

Achieving such cross-modal capability relies heavily on the availability of high-quality audio-visual data.
Recent instruction-tuning datasets~\citep{avinstruct,javisinst-und_javisgpt} are often constructed via automated pipelines that leverage the advanced reasoning capabilities of Large Language Models (LLMs) to synthesize Question-Answer (QA) pairs from intermediate video captions.
Considering MLLMs' modality bias and hallucinations during joint audio-visual understanding~\citep{avhbench}, most methods~\citep{longvale,omnivinci,longshotbench} generate separate descriptions for visual and audio modalities, where videos are split into short segments to capture more visual details. Afterwards, QA pairs are directly generated from the long, detail-heavy descriptions.
However, this paradigm exhibits three limitations:
(1) It breaks the inherent audio-visual associations within videos~\citep{egoavu};
(2) Independent segment descriptions lead to incoherent narratives. For instance, a ``person in white'' might be described as a ``person with yellow hair'' in the next segment, making it hard for LLMs to track entities;
(3) Coupling long-text comprehension and QA synthesis into a single step tends to yield questions with weaker long-term temporal and cross-modal dependencies.

To address these problems, we introduce an automated audio-visual QA generation pipeline consisting of two main stages.
First, to mitigate narrative incoherence and broken sound-source associations, we propose \textbf{Entity-Anchored Scripting}. This stage represents the video as a structured, script-like text, which includes a summary, a list of main entities, and segment-wise information with timestamps (covering speech transcriptions, non-speech sounds, and visual descriptions). The main entity list serves as a global prior to unify references across all segments. Meanwhile, we associate speech with the speaker, linking the visual and audio modalities through the same entity identifiers.
Second, to generate questions with long-term temporal spans and strong cross-modal dependencies, we design \textbf{Clue-Guided QA Generation}. Building on the entity-anchored script, we first require the LLM to identify clues centered on specific audio-visual tasks, instead of letting it generate questions from the entire text. These clues from multiple segments and modalities collectively form comprehensive reasoning chains. The model then focuses on these clues to generate QA pairs featuring long-term temporal spans and strong cross-modal dependencies.

\begin{figure}[!tbp]
    \centering
    \includegraphics[trim=0 20 290 0, clip, width=1.0\textwidth]{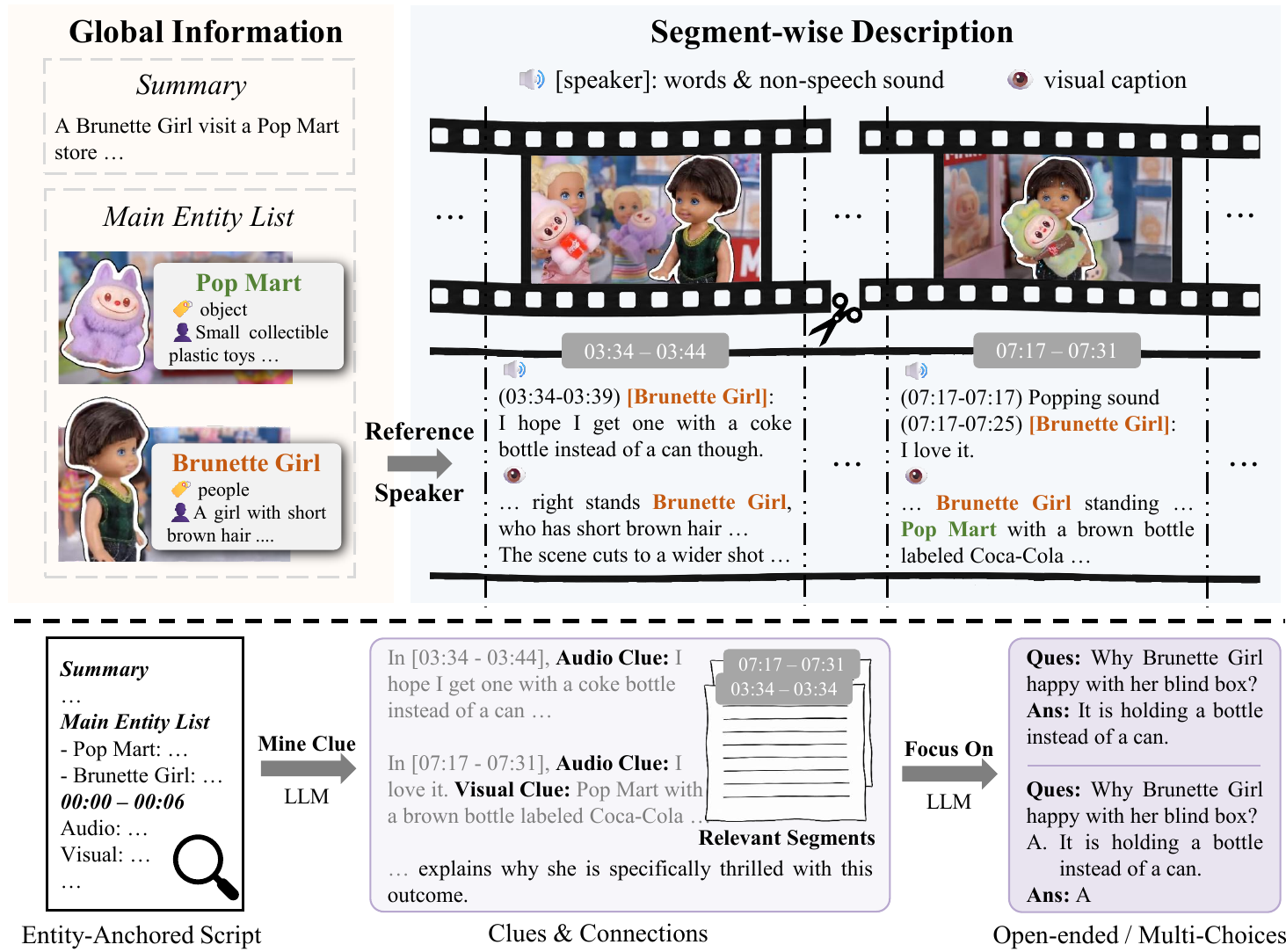}
    \caption{Overview of our pipeline.
    The pipeline first transforms audio-visual videos into entity-anchored scripts (upper part), comprising a summary, a main entity list, and structured segment-wise descriptions that integrate speech, sounds, and visual information. By utilizing consistent identifiers (represented by \textcolor{orange_fig}{orange} and \textcolor{green_fig}{green} text) from the main entity list, we ensure narrative coherence across different segments and associate speech content with visual entities. Subsequently, the clue-guided QA generation strategy (lower part) prompts LLMs to mine cross-segment and cross-modal clues from the script, followed by a locally-focused generation step to produce QA pairs based on the clues and relevant segments.
    }
    \label{fig:overview}
\end{figure}
Leveraging this pipeline, we construct OmniVideo-100K, an instruction-tuning dataset containing 100K automatically generated audio-visual QA pairs, and OmniVideo-Test, a test set with 505 human-verified samples. Both datasets cover ten diverse audio-visual tasks. 
Experimental results on OmniVideo-Test reveal that while open-source MLLMs perform relatively well in basic audio-visual semantic association, they remain constrained in fine-grained temporal alignment and cross-modal reasoning. 
Notably, after fine-tuning on OmniVideo-100K, models based on VITA-1.5~\citep{vita-1.5}, Qwen2.5-Omni-7B~\citep{qwen2.5-omni} and Qwen3-Omni-30B-A3B-Instruct~\citep{qwen3-omni} achieve performance gains of 20.59\%, 17.82\% and 13.86\% on OmniVideo-Test, respectively. Furthermore, these improvements demonstrate strong generalization across multiple existing benchmarks, including Video-MME~\citep{video-mme}, Daily-Omni~\citep{daily-omni}, and JointAVBench~\citep{jointavbench}, among others. 
Qualitative analysis confirms that fine-tuned models effectively capture audio-visual clues, revealing a clear shift from a reliance on single-modality perception to cross-modal synergy.

\section{Method}
In this section, we detail our pipeline, as illustrated in Fig.~\ref{fig:overview}. Our framework is organized into two primary modules: Entity-Anchored Video Scripting (Sec.~\ref{sec:script}) and Clue-Guided QA Generation (Sec.~\ref{sec:qa}). Prompts are provided in Sec.~\ref{sec:prompts}. Finally, we present an overview of our instruction-tuning dataset, OmniVideo-100K, and the test set, OmniVideo-Test (Sec.~\ref{sec:dataset}).

\subsection{Entity-Anchored Video Scripting}
\label{sec:script}
We leverage MLLMs to transform audio-visual videos into structured scripts. A script consists of a summary, a main entity list, and a body of sequential segments. As shown in Fig.~\ref{fig:overview} (blue region), each segment includes its timestamp, visual description, and audio information (chronologically ordered speaker-labeled speech and non-speech sounds).
We integrate discrete segments into a coherent script through unified entity identifiers across segment descriptions. Simultaneously, we tag speech with its speaker labels to associate audio and visual information.

\textbf{Main Entity List.}
Before modality decoupling and video segmentation, we use MLLMs to identify main active entities (e.g., people, animals, and objects) that are highly relevant to the video content. For each entity, we generate a unique descriptive identifier (e.g., ``Brunette Girl'') along with a detailed feature description. These identifiers and descriptions collectively form the main entity list, which serves as a global prior to guide and constrain the subsequent scripting process.

\textbf{Audio Information with Timestamps.}
We process the audio stream extracted from the raw video. We use MLLMs to generate transcriptions with start and end timestamps, where segmentation is determined by natural pauses or semantic completeness. Meanwhile, the model also identifies common non-speech sounds and provides music descriptions, all with timestamps.

\textbf{Coherent Segmented Visual Narratives.}
We first partition the video into primary segments to establish a temporal backbone for the script, setting a target duration (15 seconds in our implementation) to capture more visual details. We use speech timestamps as boundaries to prevent semantic gaps. Neighboring short segments are merged up to the target duration to improve efficiency. For videos without speech, the entire video is treated as a single primary segment.

Building upon this temporal backbone, we generate visual captions for each primary segment based on video frames. The main entity list is used as a prior to ensure consistent entity references across all descriptions. Specifically, if a primary segment exceeds the target duration, we split it evenly into multiple smaller sub-segments for sequential processing. The final visual caption for the primary segment is synthesized by concatenating the timestamps and corresponding descriptions of its constituent sub-segments.

For each individual segment (either a primary segment or a sub-segment), the MLLM analyzes every shot or significant scene change across four dimensions: Setting\&Environment, Characters\&Objects, Actions\&Interactions, and Cinematography. The observed content is then synthesized into coherent paragraphs, each corresponding to a shot or scene.

\textbf{Sound-Source Association.}
We use MLLMs to identify the speaker for each transcription, which is essential for associating speech with visual narratives in multi-person scenarios. We input the raw video, transcriptions, and the main entity list into the MLLM, requiring it to assign speaker labels by jointly analyzing visual and audio features. This process can also handle off-screen speakers, such as narrators, who may not have a visual presence. If a speaker is already listed in the main entity list, we use the existing identifier; otherwise, a unique identifier is generated for the new speaker.

To provide a broader narrative context, we also generate a summary of the raw video, with entity references constrained by the main entity list. Finally, all processed information is integrated into a complete audio-visual script.

\subsection{Clue-Guided QA Generation}
\label{sec:qa}
Based on the coherent audio-visual script, we adopt a clue-guided strategy to construct QA pairs with long-term temporal spans and deep cross-modal dependencies. This strategy is divided into two steps: Global Clue Mining and Locally Focused Generation.

\textbf{Step 1: Global Clue Mining.}
We use LLMs to perform a scan of the full script to extract clues needed for task-specific QA generation. This process emphasizes the integration of information across multiple segments and modalities.
For instance, in causal reasoning, we require the model to retrieve and link multimodal information distributed throughout the script to construct multiple causal chains. The true cause of an event can often only be understood by combining audio cues from some segments with the visual content of other segments.

For each identified chain, the model provides the relevant segment timestamps and a logical description of the audio-visual synergy. Compared to direct QA generation, this process converts the model's implicit understanding of complex narratives into explicit, step-by-step reasoning.

\textbf{Step 2: Locally Focused Generation.}
After identifying high-value clue chains, we move to the QA generation phase. We use the logical descriptions and segment timestamps from the previous step as contextual prompts to guide the model to focus on the key segments during QA construction. This focused approach filters out irrelevant content from the full script, reducing the model's cognitive load. Most importantly, since the generation is anchored on pre-verified clues, this strategy helps ensure that the resulting QA pairs meet the requirements for both long-term temporal spans and audio-visual synergy.

\subsection{Dataset Overview}
\label{sec:dataset}
\textbf{Task Taxonomy.}
We define ten general audio-visual QA tasks (detailed in Sec.~\ref{sec:tasks}) organized into a three-level cognitive framework. This framework ranges from basic perception to advanced reasoning, ensuring the comprehensiveness of our data:
\begin{itemize}
    \item \textbf{Alignment:}
    This level includes Fine-Grained Perception and Scene Transformation Detection.
    As the foundation of the framework, it focuses on the ability to perceive and synchronize audio and visual information along the timeline.
    \item \textbf{Understanding:}
    This level covers Context Understanding, Comparison, Sentiment Analysis, Event Sequence Ordering and Summarization.
    It emphasizes semantic analysis, requiring the model to link cross-modal information and grasp the narrative content.
    \item \textbf{Reasoning:}
    This level comprises Causal Reasoning, Future Prediction and Hypothetical Reasoning.
    Representing advanced cognitive abilities, it requires the model to go beyond surface-level understanding to perform logical deduction and abstract thinking.
\end{itemize}

\textbf{Video Curation.}
We collect videos from online video platforms. Based on the video categorization in OmniVideoBench~\citep{omnivideobench}, we initiate a search keyword pool comprising seven labels: vlog, news, cartoon, sports, documentary, tv, and ego. We iteratively expand this pool with newly collected video tags to ensure diversity. We filter out low-quality videos with resolutions below 480p and retain only English videos. 
Following~\citet{finevideo}, we further filter videos based on visual dynamics and word density to ensure rich audio-visual information. Furthermore, to prevent models from relying on on-screen text rather than audio, we employ the video-subtitle-extractor tool\footnote{https://github.com/YaoFANGUK/video-subtitle-extractor} to exclude videos containing hard-coded subtitles.

\textbf{Dataset Composition.}
Based on the task taxonomy and curated videos, we construct datasets for training and evaluation using our pipeline.
The models employed in this process include
Gemini-2.5-Pro~\cite{gemini2.5} and Gemini-3-Pro~\cite{gemini3}.
% During this process, we employ advanced closed-source MLLMs to ensure the high quality and diversity of the generated data.
It is worth noting that since most queries in basic Alignment and Context Understanding tasks often target localized events
within short clips, we directly prompt the model to mine clues and generate QA pairs in a single pass, rather than utilizing the full multi-step strategy.

\begin{figure}[!tbp]
    \centering
    \begin{subfigure}[b]{0.5\textwidth}
        \centering
        \includegraphics[width=\textwidth, trim=230 230 230 230, clip]{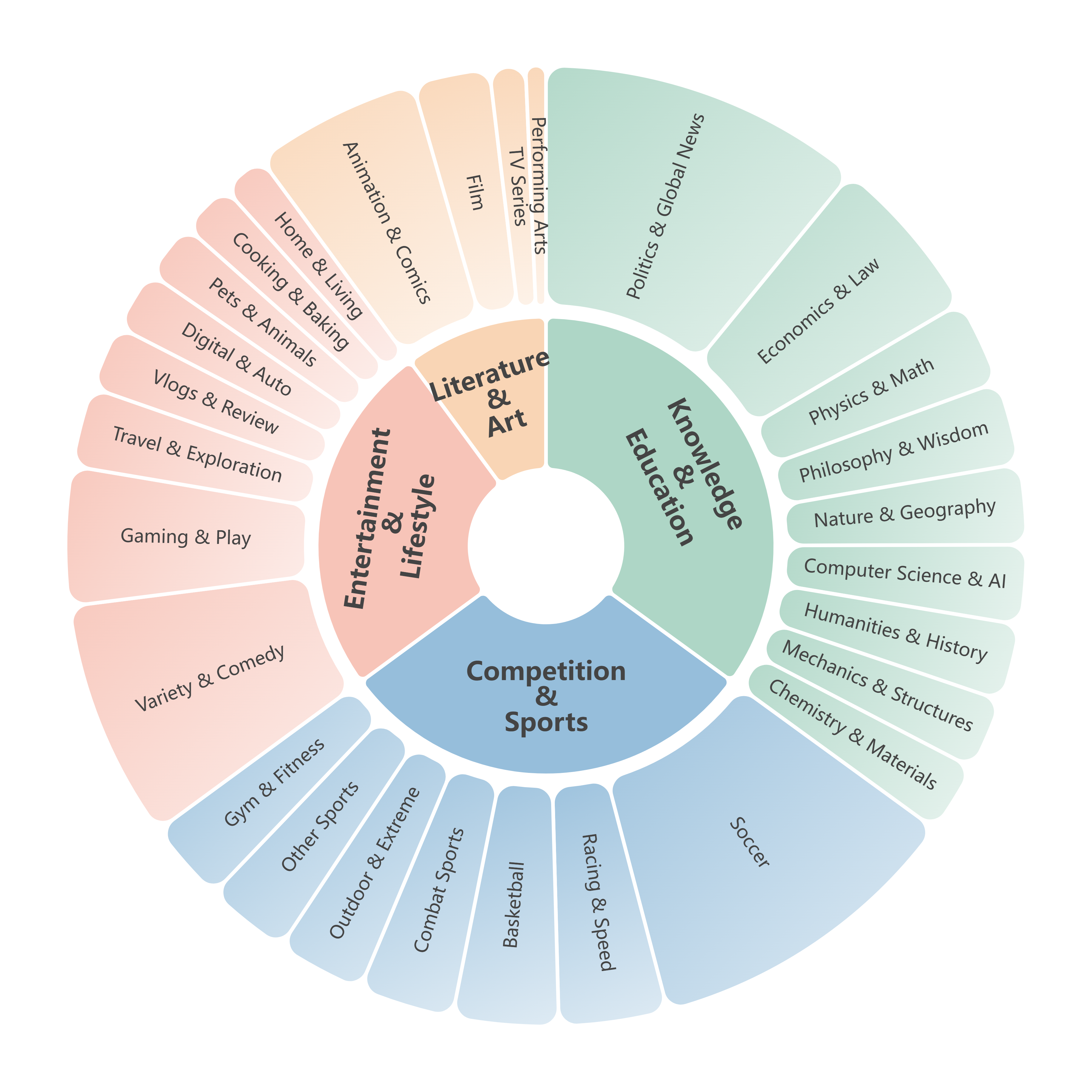} 
        \caption{Video category distribution.}
        \label{fig:category}
    \end{subfigure}
    \hfill
    \begin{subfigure}[b]{0.48\textwidth}
        \begin{subfigure}[b]{\textwidth}
            \centering
            \includegraphics[width=\textwidth]{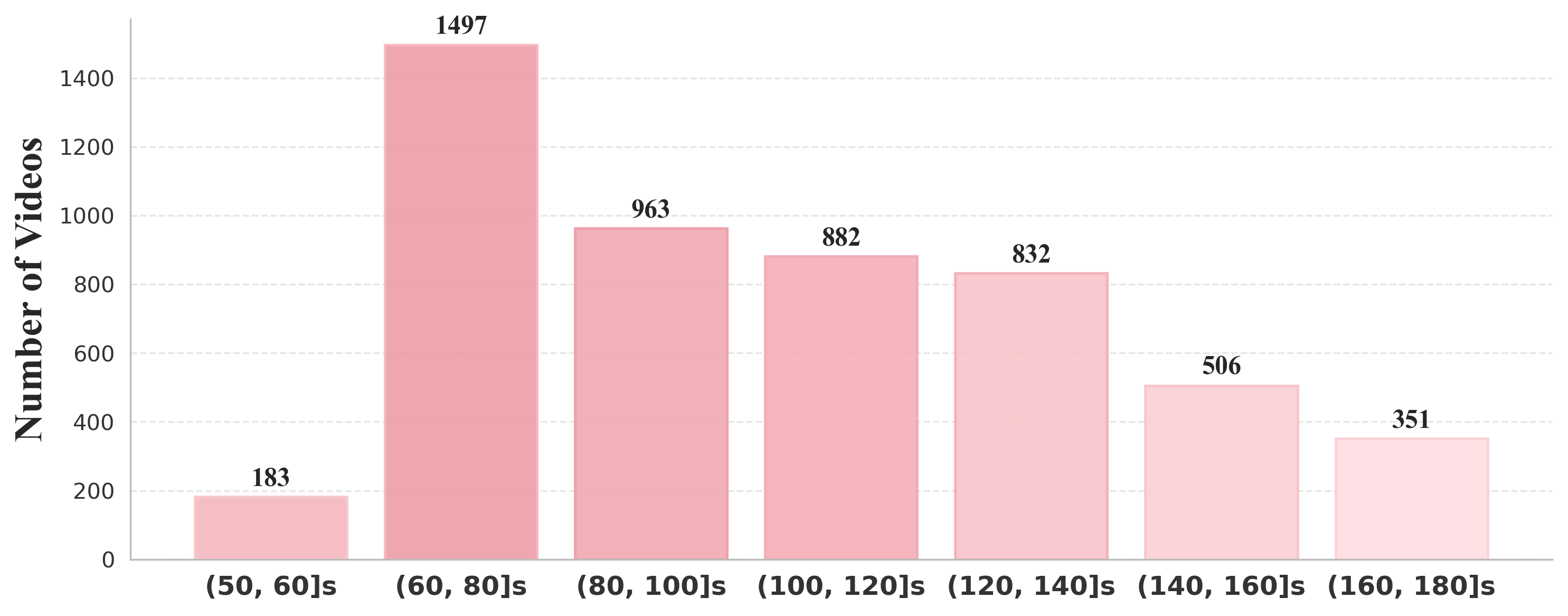}
            \caption{Video length statistics.}
            \label{fig:length}
        \end{subfigure}
        \begin{subfigure}[b]{\textwidth}
            \centering
            \includegraphics[width=\textwidth]{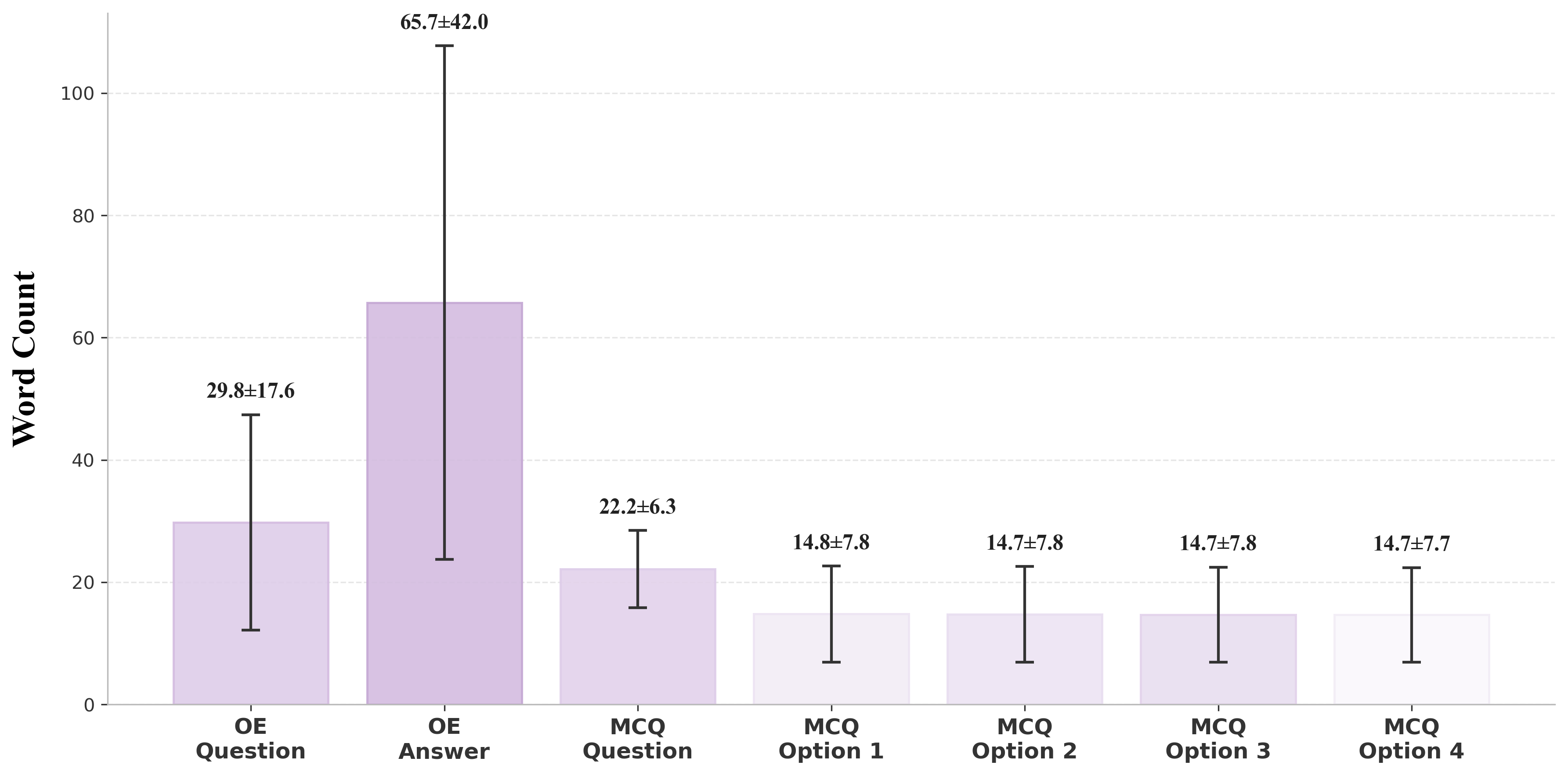}
            \caption{Word count statistics (mean±std) for Open-Ended and Multiple-Choice samples.}
            \label{fig:word}
        \end{subfigure}
    \end{subfigure}
    \caption{Statistics of the OmniVideo-100K dataset. (a) Video category distribution, illustrating a diverse range of real-world domains. (b) Video length distribution, showing that the content predominantly covers a range of 1--3 minutes. (c) Word count statistics (mean±std). The notably higher word count in Open-Ended (OE) answers stems from the inclusion of detailed explanations alongside final conclusions. Meanwhile, the lengths of the four Multiple-Choice (MCQ) options are balanced to mitigate potential length bias.}
    \label{fig:statistics}
\end{figure}

The instruction-tuning set, OmniVideo-100K, comprises 100K automatically generated QA pairs from 5214 videos. The distributions of video categories and lengths are illustrated in Fig.~\ref{fig:category} and Fig.~\ref{fig:length}, respectively. The samples consist of both Open-Ended (OE) and Multiple-Choice Questions (MCQ) in a 7:3 ratio, with samples evenly distributed across all ten audio-visual tasks.
We further analyzed the word count statistics in Fig.~\ref{fig:word}. Notably, the OE answers are significantly longer than the MCQ options, as they provide a detailed explanation alongside the final conclusion. The lengths of the four MCQ options are carefully balanced to prevent models from exploiting length bias as a shortcut for learning.

The corresponding test set, OmniVideo-Test, contains 505 multiple-choice QA pairs from 264 videos. These samples are initially generated by our pipeline and then undergo manual review. Each QA pair is verified for factual accuracy, cross-modal dependency, and answer uniqueness. Specifically, we discarded questions that could be answered or easily guessed using only a single modality. Additionally, any pairs with factual errors or ambiguous options were removed. This filtering process results in a final pass rate of approximately 38.14\%.

\begin{table}[!htbp]
\centering
\caption{Comparison of OmniVideo-100K with existing audio-visual video datasets for instruction tuning. We highlight that our dataset provides complex temporal tasks, evidence-based QA, and structured narratives that can be used for broader application. * denotes the automated pipeline leverages annotations from other datasets rather than starting from raw video.}
\resizebox{\linewidth}{!}{
  \begin{tabular}{l|ccccccc}
    \toprule
    \textbf{Datasets} & \textbf{\# Samples} & \textbf{Domain} & \textbf{Avg. Length} & \textbf{Annotation} & \textbf{Complex Temporal} & \textbf{Evidence Chain} & \textbf{Structured Narratives}   \\
    \midrule
    AVSD~\citep{avsd} & 8K & Open & 30s  & Manual & \ding{55} & \ding{55} & \ding{55} \\
    Pano-AVQA~\citep{pano-avqa} & 42.8K & Panoramic & 5s & Manual & \ding{55} & \ding{55} & \ding{55} \\
    Music-AVQA~\citep{music-avqa} & 32K & Music & 60s & Manual & \ding{55} & \ding{55} & \ding{55} \\
    AVQA~\citep{avqa} & 40K & Open & 10s & Manual & \ding{55} & \ding{55} & \ding{55} \\
    \midrule
    JavisInst-Und~\citep{javisinst-und_javisgpt} & 110K & Open & 10s & Automatic\rlap{*} & \ding{55} & \ding{55} & \ding{55} \\
    EgoAVU-Instruct~\citep{egoavu} & 3M & Ego & 1--6min & Automatic & \ding{55} & \ding{55} & \ding{55} \\
    \midrule
    \textbf{OmniVideo-100K} & 100K & Open & 103s & Automatic & \ding{52} & \ding{52} & \ding{52} \\
    \bottomrule
  \end{tabular}
  \label{tab:compare}
}
\end{table}

\textbf{Comparison with Existing Datasets.}
We compare OmniVideo-100K with several existing datasets, as summarized in Table~\ref{tab:compare}. Early datasets, such as Music-AVQA~\citep{music-avqa} and AVQA~\citep{avqa}, rely on manual annotations, which limits scalability. While JavisInst-Und~\citep{javisinst-und_javisgpt} employs an automated pipeline to expand the dataset size, it remains dependent on pre-existing annotations and its video durations are relatively short. In contrast, EgoAVU-Instruct~\citep{egoavu} covers longer video spans, but it is restricted to the egocentric domain.

Furthermore, whereas previous datasets mostly focus on simple temporal relationships, such as answering before/after queries of audio and visual events, OmniVideo-100K introduces more challenging objectives including fine-grained alignment and ordering tasks.
% 我们线索引导QA生成策略将问答对锚定在具体的视听线索上，不同于以往数据集对于因果等推理只给出简短结论，我们的开放式答案还包含了基于视听线索的推理解释。除此之外，结构化script作为我们的附加产物，其将视频视听信息进行了解耦与分段对齐，不仅为QA的生成提供了支撑，也可以为视频编辑、生成等任务提供极具价值的中间表示。
Meanwhile, our clue-guided strategy anchors QA pairs to explicit audio-visual evidence chains. Consequently, our open-ended answers include detailed reasoning explanations grounded in cross-modal clues, rather than only the brief conclusions commonly found in prior datasets. Beyond these high-quality QA pairs, the scripts produced by our pipeline serve as a valuable byproduct of OmniVideo-100K. Compared with single-paragraph narratives that entangle audio and visual modalities, our structured scripts not only support QA generation but also provide reusable intermediate representations for broader applications like video editing.

\section{Experiments}
To verify the effectiveness of OmniVideo-100K, we perform full-parameter fine-tuning on VITA-1.5, Qwen2.5-Omni-7B and Qwen3-Omni-30B-A3B-Instruct. We employ LLaMA-Factory~\citep{llamafactory} for the Qwen-series models, while VITA-1.5 is fine-tuned using its official implementation. Detailed parameter settings are shown in Table~\ref{tab:args}.

\subsection{Main Results}
We evaluate and compare the performance of the fine-tuned MLLMs, the baseline models, and several other open-source models (video-SALMONN 2+~\citep{video-salmonn2}, OmniVinci~\citep{omnivinci}, Uni-MoE-2.0-Omni~\citep{uni-moe-2.0-omni}, and MiniCPM-o 4.5~\citep{minicpm-o4.5}) on both OmniVideo-Test and multiple existing benchmarks. During inference, we sample 16 frames for VITA-1.5 and 64 for other models.

\subsubsection{Performance on OmniVideo-Test}
\begin{table}[!htbp]
\centering
\caption{Performance comparison on OmniVideo-Test. We report results across task dimensions and video durations. Red values denote the performance gains achieved after fine-tuning on the OmniVideo-100K dataset.}
\label{tab:avs-bench_results}
\resizebox{\linewidth}{!}{
\begin{tabular}{lc|w{c}{2cm}|ccw{c}{2cm}|cw{c}{2cm}}
\toprule
\textbf{Models} & \textbf{Size} & \textbf{Overall} & \textbf{Alignment} & \textbf{Understanding} & \textbf{Reasoning} & \textbf{(0, 2]min} & \textbf{(2, 5]min}\\
\midrule
Human & - & 100 & - & - & - & - & -\\
Gemini-3.1-Pro & - & 83.96 & 83.62 & 84.50 & 83.21 & 82.61 & 84.59 \\
\midrule
% \multicolumn{8}{c}{\textit{Open Source MLLMs}} \\ 
% \midrule
VITA-1.5 & 7B & 40.99 & 28.45 & 43.02 & 48.09 & 44.72 & 39.24 \\
Qwen2.5-Omni & 7B & 42.77 & 39.66 & 46.51 & 38.17 & 38.51 & 44.77 \\
video-SALMONN 2+ & 7B & 45.15 & 37.93 & 49.22 & 43.51 & 46.58 & 44.48 \\
OmniVinci & 9B & 47.13 & 43.97 & 51.55 & 41.22 & 47.20 & 47.09 \\
MiniCPM-o 4.5 & 9B & 55.25 & 56.90 & 58.14 & 48.09 & 53.42 & 56.10 \\
Qwen3-Omni & 30B & 49.70 & 43.10 & 55.04 & 45.04 & 49.07 & 50.00 \\
Uni-MoE-2.0-Omni & 30B & 46.93 & 39.66 & 52.71 & 41.98 & 46.58 & 47.09 \\
\midrule
% \multicolumn{8}{c}{\textit{MLLMs trained with OmniVideo-100K}} \\ 
% \midrule
OmniVideo-7B\textsubscript{(VITA-1.5)} & 7B & 61.58\rlap{\textsubscript{\color{red}+20.59}} & 59.48\rlap{\textsubscript{\color{red}+31.03}} & 63.18\rlap{\textsubscript{\color{red}+20.16}} & 60.31\rlap{\textsubscript{\color{red}+12.22}} & 59.01\rlap{\textsubscript{\color{red}+14.29}} & 62.79\rlap{\textsubscript{\color{red}+23.55}} \\
OmniVideo-7B\textsubscript{(Qwen2.5-Omni)} & 7B & 60.59\rlap{\textsubscript{\color{red}+17.82}} & 62.93\rlap{\textsubscript{\color{red}+23.27}} & 62.40\rlap{\textsubscript{\color{red}+15.89}} & 54.96\rlap{\textsubscript{\color{red}+16.79}} & 54.66\rlap{\textsubscript{\color{red}+16.15}} & 63.37\rlap{\textsubscript{\color{red}+18.60}} \\
OmniVideo-30B\textsubscript{(Qwen3-Omni)} & 30B & 63.56\rlap{\textsubscript{\color{red}+13.86}} & 60.34\rlap{\textsubscript{\color{red}+17.24}} & 67.05\rlap{\textsubscript{\color{red}+12.01}} & 59.54\rlap{\textsubscript{\color{red}+14.50}} & 62.11\rlap{\textsubscript{\color{red}+13.04}} & 64.24\rlap{\textsubscript{\color{red}+14.24}} \\
% \midrule
% \multicolumn{8}{c}{\textit{Audio-Only}} \\ 
% \midrule
% MiniCPM-o 4.5 & 9B & 45.74 & 36.21 & 51.94 & 41.98 & 45.34 & 45.93 \\
% Qwen3-Omni & 30B & 46.14 & 37.07 & 52.71 & 41.22 & 48.45 & 45.06 \\
% \midrule
% \multicolumn{8}{c}{\textit{Visual-Only}} \\ 
% \midrule
% MiniCPM-o 4.5 & 9B & 47.92 & 42.24 & 50.00 & 48.85 & 51.55 & 46.22 \\
% Qwen3-Omni & 30B & 47.13 & 42.24 & 48.84 & 48.09 & 45.96 & 47.67 \\
\bottomrule
\end{tabular}
}
\end{table}
\textbf{Performance Discrepancies Across Tasks.}
Table~\ref{tab:avs-bench_results} shows the performance on OmniVideo-Test (refer to Table~\ref{tab:avs-bench_results_details} for detailed per-task results). Existing models generally perform better on the \textbf{Understanding} tasks than on Alignment and Reasoning tasks. For example, video-SALMONN 2+ achieves 49.22\% on Understanding, but only 37.93\% and 43.51\% on Alignment and Reasoning, respectively. This indicates that the ability to establish preliminary semantic connections between visual and audio content and achieve basic understanding is relatively well-handled by current models.
In contrast, for the \textbf{Alignment} tasks, which involve Fine-Grained Perception and Scene Transformation Detection, the main difficulty lies in precisely aligning visual and audio information in the temporal dimension.
Except for MiniCPM-o 4.5, most open-source models perform similarly, reflecting that existing models still have certain limitations in temporal alignment. For the \textbf{Reasoning} tasks, models not only need cross-modal semantic understanding but must integrate multimodal cues to perform logical reasoning and derive abstract conclusions.
However, even for MiniCPM-o 4.5, which exhibits strong results in the first two dimensions, its accuracy in this dimension still does not exceed 50\%, reflecting that current open-source MLLMs still have significant room for improvement in deep cross-modal reasoning.

\begin{figure}[!tbp]
    \centering
    \includegraphics[trim=5 95 165 5, clip, width=1.0\textwidth]{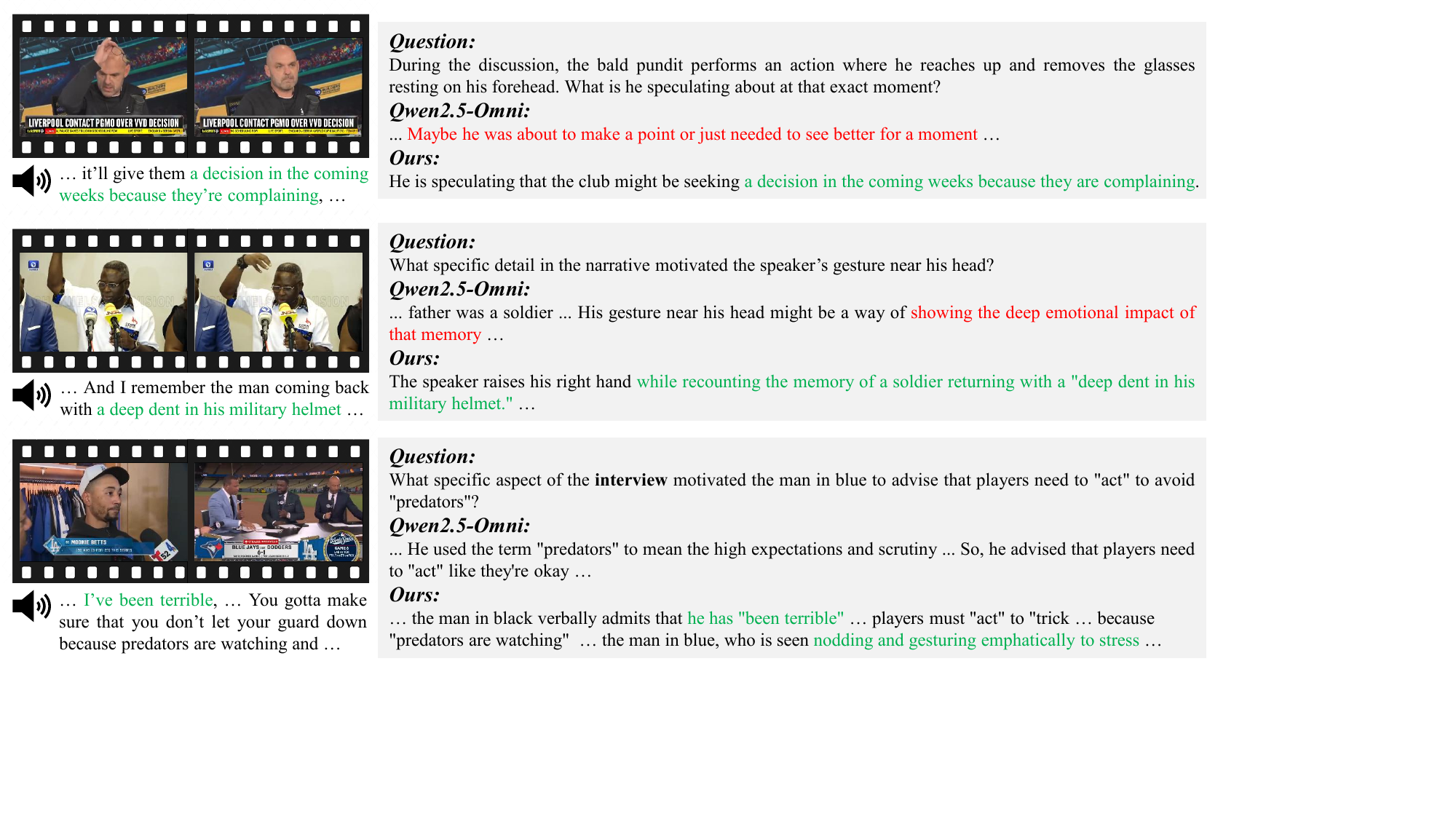}
    \caption{
    Qualitative comparison between the baseline (Qwen2.5-Omni-7B) and our fine-tuned model on OmniVideo-Test. We directly input the questions into the models to obtain open-ended responses. Green text highlights accurate answers grounded in specific evidence, while red text indicates vague speculation or erroneous content. Compared to the baseline, our model reduces reliance on unimodal speculation (Case 1), exhibits improved fine-grained temporal alignment (Case 2), and facilitates cross-temporal synergistic reasoning (Case 3).}
    \label{fig:avsbench}
\end{figure}
\textbf{Improvements from OmniVideo-100K.} Models fine-tuned on OmniVideo-100K exhibit significant performance gains of 20.59\%, 17.82\% and 13.86\% over their respective baselines, effectively enhancing synergetic understanding in audio-visual scenarios. Fig.~\ref{fig:avsbench} presents a qualitative comparison on OmniVideo-Test between the baseline Qwen2.5-Omni and its fine-tuned counterpart. To provide a more intuitive analysis, we conduct tests using the open-ended form. We find that the baseline often relies on single-modality cues to generate speculative responses, rather than grounding its answers in the actual audio-visual content (e.g., Case 1). In Case 2, the baseline only captures audio content (``father was a soldier'') proximate to the target timestamp, thereby producing an erroneous response which highlights limitations in fine-grained temporal alignment. Furthermore, in Case 3, the baseline fails due to its inability to associate information from the preceding segment. In contrast, the fine-tuned model successfully leverages both cross-temporal and cross-modal clues to complete the reasoning. It not only incorporates content from the earlier interview but also attends to the visual actions of the man in blue. These qualitative results demonstrate that OmniVideo-100K not only improves temporal alignment but also facilitates a transition from single-modality reliance to joint multimodal understanding.

\begin{table}[!htbp]
\centering
\caption{Performance comparison under different modality configurations on OmniVideo-Test. We evaluate two open-source models across audio-visual, audio-only, and visual-only settings to analyze cross-modal dependency. Green and red values denote performance changes relative to the full audio-visual setting.}
\label{tab:avs-bench_results_modal}
\resizebox{\linewidth}{!}{
\begin{tabular}{lc|w{c}{1.8cm}|ccw{c}{2cm}|cw{c}{2cm}}
\toprule
\textbf{Models} & \textbf{Size} & \textbf{Overall} & \textbf{Alignment} & \textbf{Understanding} & \textbf{Reasoning} & \textbf{(0, 2]min} & \textbf{(2, 5]min}\\
\midrule
\rowcolor{gray!15}
\multicolumn{8}{c}{\textit{Audio-Visual}} \\ 
\midrule
MiniCPM-o 4.5 & 9B & 55.25 & 56.90 & 58.14 & 48.09 & 53.42 & 56.10 \\
Qwen3-Omni & 30B & 49.70 & 43.10 & 55.04 & 45.04 & 49.07 & 50.00 \\
\midrule
\rowcolor{gray!15}
\multicolumn{8}{c}{\textit{Audio-Only}} \\ 
\midrule
MiniCPM-o 4.5 & 9B & 45.74\rlap{\textsubscript{\color{green}-9.51}} & 36.21\rlap{\textsubscript{\color{green}-20.69}} & 51.94\rlap{\textsubscript{\color{green}-6.20}} & 41.98\rlap{\textsubscript{\color{green}-6.11}} & 45.34\rlap{\textsubscript{\color{green}-8.08}} & 45.93\rlap{\textsubscript{\color{green}-10.17}} \\
Qwen3-Omni & 30B & 46.14\rlap{\textsubscript{\color{green}-3.56}} & 37.07\rlap{\textsubscript{\color{green}-6.03}} & 52.71\rlap{\textsubscript{\color{green}-2.33}} & 41.22\rlap{\textsubscript{\color{green}-3.82}} & 48.45\rlap{\textsubscript{\color{green}-0.62}} & 45.06\rlap{\textsubscript{\color{green}-4.94}} \\
\midrule
\rowcolor{gray!15}
\multicolumn{8}{c}{\textit{Visual-Only}} \\ 
\midrule
MiniCPM-o 4.5 & 9B & 47.92\rlap{\textsubscript{\color{green}-7.33}} & 42.24\rlap{\textsubscript{\color{green}-14.66}} & 50.00\rlap{\textsubscript{\color{green}-8.14}} & 48.85\rlap{\textsubscript{\color{red}+0.76}} & 51.55\rlap{\textsubscript{\color{green}-1.87}} & 46.22\rlap{\textsubscript{\color{green}-9.88}} \\
Qwen3-Omni & 30B & 47.13\rlap{\textsubscript{\color{green}-2.57}} & 42.24\rlap{\textsubscript{\color{green}-0.86}} & 48.84\rlap{\textsubscript{\color{green}-6.20}} & 48.09\rlap{\textsubscript{\color{red}+3.05}} & 45.96\rlap{\textsubscript{\color{green}-3.11}} & 47.67\rlap{\textsubscript{\color{green}-2.33}} \\
\bottomrule
\end{tabular}
}
\end{table}

\textbf{Cross-modal Dependency Analysis.}
% 基于表~\ref{tab:avs-bench_results}的结果，我们选择了两个代表性的开源模型：总分最高的MiniCPM-o 4.5和参数规模较大的Qwen3-Omni-30B-A3B-Instruct，以进一步分析OmniVideo-Test的跨模态依赖性。我们将它们在仅音频和仅视觉设置下的表现与全模态进行了对比。如表~\ref{tab:avs-bench_results_modal} 所示，两个模型在受限于单一模态时性能均有所下降，这证实了我们测试集中的样本确实需要跨模态协同。值得注意的是，尽管MiniCPM-o 4.5和Qwen3-Omni在单模态设置下的表现相近（差距均不到1%），但MiniCPM-o 4.5在切换到全模态时提升了7--9\%，而Qwen3-Omni仅提升了3\%左右，获得了远超后者的协同增益。
Based on the results in Table~\ref{tab:avs-bench_results}, we select two representative open-source models, namely MiniCPM-o 4.5, which achieves the highest overall performance among open-source models, and Qwen3-Omni, which has a larger parameter scale, to further analyze the cross-modal dependency of OmniVideo-Test. We compare their performance in audio-only and visual-only settings against the audio-visual mode. As shown in Table~\ref{tab:avs-bench_results_modal}, both models exhibit a performance decline when limited to a single modality, demonstrating that the samples in our test set require cross-modal synergy. Notably, although MiniCPM-o 4.5 and Qwen3-Omni show comparable unimodal performance with gaps below 1\%, MiniCPM-o 4.5 gains a larger increase when both modalities are provided. Specifically, MiniCPM-o 4.5 obtains an improvement of 7\% to 9\% while Qwen3-Omni improves by only about 3\%, indicating a higher synergistic gain.

\subsubsection{Generalization on Existing Benchmarks}
\begin{table}[!htbp]
\centering
\caption{
Performance comparison on audio-visual and general video benchmarks. Due to the 5-minute audio truncation limit of Qwen2.5-Omni, results for Video-MME-v2, OmniVideoBench, JointAVBench, and FutureOmni are reported only for samples within this duration; the row ``\# Samples'' denotes the number of samples retained. Red values denote the performance gains achieved by fine-tuned models.
}
\label{tab:existing_bench_results}
\resizebox{\linewidth}{!}{
\begin{tabular}{lc|cc|cccc}
\toprule
\textbf{Models} & \textbf{Size} & \textbf{Video-MME}\textsubscript{short} & \textbf{Video-MME-v2} & \textbf{Daily-Omni} & \textbf{OmniVideoBench} & \textbf{JointAVBench} & \textbf{FutureOmni} \\
\midrule
\rowcolor{gray!15}
\# Samples & - & 900 & 328 & 1197 & 509 & 2153 & 960 \\
\midrule
VITA-1.5 & 7B & 70.63 & 5.95 & 52.63 & 36.35 & 44.77 & 48.65 \\
Qwen2.5-Omni & 7B & 75.56 & 10.28 & 62.41 & 36.54 & 54.44 & 48.85 \\
video-SALMONN 2+ & 7B & 74.22 & 11.50 & 62.57 & 35.36 & 56.39 & 51.88 \\
OmniVinci & 9B & 77.56 & 11.59 & 61.32 & 36.74 & 57.55 & 52.81 \\
MiniCPM-o 4.5 & 9B & 83.22 & 14.48 & 80.20 & 41.06 & 55.18 & 52.29 \\
Qwen3-Omni & 30B & 82.00 & 14.31 & 74.27 & 43.84 & 63.17 & 53.44 \\
Uni-MoE-2.0-Omni & 30B & 78.89 & 8.82 & 64.33 & 38.55 & 57.55 & 52.81 \\
\midrule
OmniVideo-7B\textsubscript{(VITA-1.5)} & 7B & 67.67\rlap{\textsubscript{\color{green}-2.96}} & 
7.42\rlap{\textsubscript{\color{red}+1.47}} & 55.39\rlap{\textsubscript{\color{red}+2.76}} & 36.94\rlap{\textsubscript{\color{red}+0.59}} & 57.41\rlap{\textsubscript{\color{red}+12.64}} & 56.35\rlap{\textsubscript{\color{red}+7.70}} \\
OmniVideo-7B\textsubscript{(Qwen2.5-Omni)} & 7B & 76.33\rlap{\textsubscript{\color{red}+0.77}} & 
8.50\rlap{\textsubscript{\color{green}-1.78}} & 69.84\rlap{\textsubscript{\color{red}+7.43}} & 39.88\rlap{\textsubscript{\color{red}+3.34}} & 60.75\rlap{\textsubscript{\color{red}+6.31}} & 55.00\rlap{\textsubscript{\color{red}+6.15}} \\
OmniVideo-30B\textsubscript{(Qwen3-Omni)} & 30B & 83.56\rlap{\textsubscript{\color{red}+1.56}} & 15.33\rlap{\textsubscript{\color{red}+1.02}} & 76.61\rlap{\textsubscript{\color{red}+2.34}} & 44.81\rlap{\textsubscript{\color{red}+0.97}} & 66.37\rlap{\textsubscript{\color{red}+3.20}} & 57.60\rlap{\textsubscript{\color{red}+4.16}} \\
\bottomrule
\end{tabular}
}
\end{table}
To evaluate the generalization of the fine-tuned models, we compare their performance across several established audio-visual benchmarks. As shown in Table~\ref{tab:existing_bench_results}, our fine-tuned models achieve consistent performance gains across Daily-Omni~\citep{daily-omni}, OmniVideoBench~\citep{omnivideobench}, JointAVBench~\citep{jointavbench}, and FutureOmni~\citep{futureomni}. VITA-1.5 demonstrates a performance increase of 0.6\%--12\% across various metrics, while Qwen2.5-Omni and the larger Qwen3-Omni achieve improvements of 3\%--7\% and 1\%--4\%, respectively.

\begin{table}[!htbp]
\centering
\caption{Performance breakdown on Daily-Omni. We compare the results of the baseline Qwen2.5-Omni against our fine-tuned model across task types.}
\label{tab:dailyomni}
\resizebox{\linewidth}{!}{
\begin{tabular}{l|cccccc}
\toprule
\textbf{Models} & \textbf{AV Event Alignment} & \textbf{Comparative} & \textbf{Context Understanding} & \textbf{Event Sequence} & \textbf{Inference} & \textbf{Reasoning} \\
\midrule
Qwen2.5-Omni & 51.26 & 68.70 & 56.48 & 57.19 & 77.27 & 75.43 \\
OmniVideo-7B\textsubscript{(Qwen2.5-Omni)} & 68.49 & 73.28 & 70.98 & 60.13 & 79.22 & 76.57 \\
\bottomrule
\end{tabular}
}
\end{table}

\textbf{Performance Breakdown.} A detailed analysis reveals both the strengths of our approach and areas for future refinement. Specifically, Qwen2.5-Omni achieves its most significant gain of 7.43\% on Daily-Omni. A more granular analysis (Table~\ref{tab:dailyomni}) reveals that this growth is primarily driven by a breakthrough in cross-modal alignment capabilities. The model achieves a remarkable 17.23\% increase in the AV Event Alignment task. However, we also observe that the overall performance gain of Qwen3-Omni on OmniVideoBench is relatively limited. To investigate the underlying cause, we break down the performance by audio types following the benchmark's settings, as shown in Table~\ref{tab:omnivideobench}.
\begin{wraptable}{r}{0.5\linewidth}
\centering
\caption{Performance breakdown on OmniVideoBench. Results are compared across audio types between the baseline Qwen3-Omni and our fine-tuned model.}
\label{tab:omnivideobench}
\resizebox{\linewidth}{!}{
\begin{tabular}{l|ccc}
\toprule
\textbf{Models} & \textbf{Music} & \textbf{Sound} & \textbf{Speech} \\
\midrule
Qwen3-Omni & 36.73 & 41.05 & 45.50 \\
OmniVideo-30B\textsubscript{(Qwen3-Omni)} & 46.94 & 36.84 & 46.59 \\
\bottomrule
\end{tabular}
}
\end{wraptable}
The results reveal that while our model improves on music and speech samples, performance on the sound category drops by 4.21\%. We attribute this decline to the scripting process, where, except for music, non-speech sounds are often represented by general, coarse-grained category descriptions to avoid over-speculation. Such descriptions can introduce noise into the QA generation process. This finding suggests that future work should integrate specialized audio models to provide more precise and granular acoustic signals, particularly for complex sounds.

\textbf{General Video Understanding.}
% In addition to specialized audio-visual benchmarks, we also report results on general video benchmarks. As shown in Table~\ref{tab:existing_bench_results}, our fine-tuned models exhibit improvements on Video-MME-v2~\citep{videommev2} (Level 1), which involves the aggregation of video frames and audio. Furthermore, the models maintain superior performance on Video-MME~\citep{video-mme} (w/ audio), demonstrating that our fine-tuning enhances cross-modal synergistic capabilities without compromising the models' original general video understanding.
In addition to specialized audio-visual benchmarks, we also evaluate our models on general video understanding tasks, specifically Video-MME~\citep{video-mme} and Video-MME-v2~\citep{videommev2}. As shown in Table~\ref{tab:existing_bench_results}, despite being fine-tuned primarily for complex cross-modal comprehension, our models maintain their original baseline performance across both general benchmarks. Although some fluctuations are observed, there is no significant performance drop; several configurations even exhibit improvements. This demonstrates that fine-tuning on OmniVideo-100K effectively enhances audio-visual synergistic capabilities without compromising the models' original general video understanding.

\subsubsection{Comparison with Existing Datasets}
% 我们选择了两个通用的开源视听数据集对qwen2.5-omni进行指令微调以进行定量比较，训练参数与OmniVideo-100K保持一致。其中，AVQA数据集为人工注释，仅包含选择题。由于一些原始视频并未找到，我们丢弃了原数据集中的243条数据，最终用于训练的有40182条数据。JavisInst-Und为自动化流水线构造的数据，共有110K条开放与多项选择混合的样本，与OmniVideo-100K相近。我们将其中Gen2Und任务相关的多轮对话中的第一轮生成视频直接丢弃，而保留剩下对话直接进行视频理解。我们同样在多个基准上测试微调后的模型，结果如表7。可以看到，在AVQA和JavisInst-Und上训练的模型在大多数基准上都发生不同程度的性能下降。
\begin{table}[!htbp]
\centering
\caption{Performance comparison of Qwen2.5-Omni fine-tuned on different audio-visual datasets across multiple benchmarks.}
\label{tab:other_datasets}
\resizebox{\linewidth}{!}{
\begin{tabular}{l|ccccccc}
\toprule
\textbf{Data} & \textbf{Video-MME}\textsubscript{short} & \textbf{Video-MME-v2} & \textbf{OmniVideo-Test} & \textbf{Daily-Omni} & \textbf{OmniVideoBench} & \textbf{JointAVBench} & \textbf{FutureOmni}\\
\midrule
Qwen2.5-Omni & 75.56 & 10.28 & 42.77 &  62.41 & 36.54 & 54.44 & 48.85 \\
\midrule
w. AVQA & 68.11 & 9.69 & 6.28 & 55.14 & 34.38 & 50.16 & 43.65 \\
w. JavisInst-Und & 59.44 & 3.22 & 38.22 & 48.96 & 32.42 & 44.36 & 58.54 \\
w. OmniVideo-100K & 76.33 & 8.50 & 60.59 & 69.84 & 39.88 & 60.75 & 55.00\\
\bottomrule
\end{tabular}
}
\end{table}
We select two open-domain audio-visual datasets, AVQA and JavisInst-Und, to fine-tune Qwen2.5-Omni for a quantitative comparison with our dataset. To ensure fairness, all training parameters are kept identical to those used for OmniVideo-100K. The AVQA dataset is manually annotated and contains only multiple-choice questions. We exclude 243 samples due to unavailable videos, resulting in a final training set of 40182 samples. JavisInst-Und is constructed via an automated pipeline and comprises 110K samples with a mix of open-ended and multiple-choice QA, which is similar in scale and format to our dataset. Note that, for this baseline, we remove the initial video-generation rounds within the multi-turn dialogues of the Gen2Und task and retain the remaining turns. As shown in Table~\ref{tab:other_datasets}, models fine-tuned on AVQA and JavisInst-Und show varying degrees of performance drops across most benchmarks compared to the original base model. In contrast, our OmniVideo-100K achieves consistent and significant gains on most benchmarks. These results show that, compared to other datasets, our dataset is more effective in enhancing the video understanding capabilities of MLLMs.

\subsection{Ablation Analysis}
\subsubsection{Clue-Guided QA Generation}
\textbf{Experimental Setup.} To investigate the impact of our Clue-Guided Generation, we compare it against a Direct Generation baseline. In this baseline, the entire video script is fed into the model, which is then prompted to directly generate QA pairs by integrating cross-segment and cross-modal clues. Since our clue-mining step produces multiple clue chains, we similarly require the baseline to output multiple QA pairs. This experiment is conducted on 20 randomly sampled videos, yielding a total of 71 QA pairs per method, which are generated by Gemini-3-Pro across all videos and encompass both Causal Reasoning and Future Prediction tasks. We ensure that the number of QA pairs per task for each video remains identical between the two methods.

\begin{wraptable}{r}{0.5\linewidth}
  \centering
  \caption{Accuracy of various MLLMs on QA pairs generated by Direct Generation and Clue-Guided Generation strategies.}
  \label{tab:wo_clue}
  \resizebox{\linewidth}{!}{
  \begin{tabular}{l|cc}
    \toprule
    \textbf{Models} & \textbf{Direct} & \textbf{Clue-Guided} \\
    \midrule
     Qwen2.5-Omni & 80.28 & 59.15 \\
     MiniCPM-o 4.5 & 78.87 & 74.65 \\
     OmniVideo-7B\textsubscript{(Qwen2.5-Omni)} & 81.69 & 80.28 \\
    \bottomrule
  \end{tabular}
  }
\end{wraptable}
\textbf{Question Difficulty.}
% As shown in Table~\ref{tab:wo_clue}, 所有模型在来自Clue-Guided的QA对上的准确率均低于其在Direct QA对上的表现，这说明Our Clue-Guided Generation strategy produces more challenging questions than the Direct Generation baseline. 此外，different models show highly similar performance on QA pairs from Direct Generation，而它们在Clue-Guided QA上的表现差异明显，这表明该策略生成的QA对能更有效地体现不同模型之间的能力差异。Qwen2.5-Omni在经过OmniVideo-100K微调后，其在已接近饱和的Direct QA上的提升相对微小，但在更具挑战性的Clue-Guided样本上，准确率从59.15\%提升至 80.28\%，表明OmniVideo-100K数据集在提升模型视听理解的有效性。
As shown in Table~\ref{tab:wo_clue}, all models achieve lower accuracy on QA pairs generated by Clue-Guided Generation than on those from the Direct Generation baseline, indicating that our strategy produces more challenging questions. Moreover, while different models exhibit highly similar performance on Direct samples, the performance gaps become more pronounced on Clue-Guided QA pairs, suggesting that our strategy better distinguishes between varying model capabilities. Notably, after fine-tuning on OmniVideo-100K, Qwen2.5-Omni shows only a marginal gain on Direct samples but a substantial improvement on the more demanding Clue-Guided set, with accuracy increasing from 59.15\% to 80.28\%. The enhancement is further corroborated by the consistent performance gains observed across multiple established benchmarks in Table~\ref{tab:existing_bench_results}, thereby verifying the effectiveness of OmniVideo-100K in enhancing audio-visual understanding.

\textbf{Temporal Scope.} We compare the average temporal span of the content covered by the questions (measured from the first evidence clue to the last) across both sets of QA pairs. For Clue-Guided QA, we directly utilize the timestamps of relevant segments from the intermediate clue chains, calculating the span by subtracting the end time of the first segment from the start time of the last. For Direct Generation, we prompt Gemini-3-Pro to determine the temporal span based on the raw video. The results reveal that Clue-Guided Generation yields an average temporal span of 144.75s, far exceeding 76.24s of the Direct Generation baseline. This demonstrates that our two-stage clue-guided approach effectively directs the model to establish connections across longer temporal distances.

\subsubsection{Main Entity List}
To verify the effectiveness of the Main Entity List in maintaining narrative coherence and mitigating sound-source mismatch, we conduct comparative experiments on 20 randomly sampled videos.

\textbf{Referential Consistency.} We develop a variant that removes the main entity list constraint during the segmented visual description phase. To assess how this referential consistency impacts QA generation, we prompt Gemini-3-Pro to generate three QA pairs per video based on the scripts. Answering these questions requires integrating entity information across multiple segments. Each pipeline generates 60 QA pairs in total. We utilize Gemini-3-Pro to judge whether the generated pairs suffer from entity confusion by referencing the raw videos. The results show that without the global prior, the error rate regarding entity confusion is 36.7\%, whereas the full pipeline reduces this error rate to 23.4\%. This indicates that, compared to inconsistent referencing across segments, the constraints from the main entity list facilitate the construction of cohesive narrative logic, which ultimately supports the generation of more reliable QA pairs.

\textbf{Audio-Visual Association.} Similarly, we construct a variant by removing speaker labels to investigate their role in cross-modal grounding. We require the model to generate three QA pairs per video based on the scripts, necessitating the correct association of speech content with the visual information of the corresponding speaker. Both the full script and the label-free script pipelines produce 60 QA pairs each. Evaluation by Gemini-3-Pro against the raw videos reveals that the audio-visual mismatch rate is 20\% without speaker labels, compared to only 10\% when labels are included. This demonstrates that explicit speaker labeling within the script effectively helps the model establish more accurate audio-visual associations, thereby enhancing the validity of the resulting QA pairs.

\subsubsection{Video-Based vs. Script-Based}
\textbf{Experimental Setup.} To evaluate the necessity of the intermediate script representation, we randomly sample 20 videos to compare QA pairs generated from raw video with those generated using the script. Each pipeline utilizes the clue-guided strategy to produce a total of 60 QA pairs across Causal Reasoning and Future Prediction tasks.

\begin{wraptable}[26]{r}{0.5\linewidth}
  \centering
  \caption{Average temporal span of QA pairs generated by video-based and script-based clue-guided pipelines across different video durations. The row “\# Samples” denotes the number of QA pairs in each duration group.}
  \label{tab:time}
  \small
  \resizebox{\linewidth}{!}{
  \begin{tabular}{l|c|cc}
    \toprule
    \textbf{Pipelines} & \textbf{Overall} & \textbf{[1,3]min} & \textbf{[8,14]min} \\
    \midrule
    \rowcolor{gray!15}
    \# Samples & 60 & 28 & 32 \\
    \midrule
    Video + Clue-Guided & 131.57s & 53.43s & 190.85s \\
    Script + Clue-Guided & 169.45s  & 53.96s & 270.50s \\
    \bottomrule
  \end{tabular}
  }
  \vspace{1.5em}
  \caption{Model accuracy on 8--14 minute QA pairs generated by video-based and script-based clue-guided pipelines under full audio-visual input and single-modality ablation settings. Green values denote the performance drop relative to the full audio-visual setting.}
  \label{tab:modal}
  \small
  \resizebox{\linewidth}{!}{
  \begin{tabular}{l|cc}
    \toprule
    \textbf{Models} & \textbf{Video + Clue-Guided} & \textbf{Script + Clue-Guided} \\
    \midrule
    \rowcolor{gray!15}
    \multicolumn{3}{c}{\textit{Audio-Visual}} \\
    \midrule
    OmniVinci & 78.12 & 68.75 \\
    MiniCPM-o 4.5 & 84.38 & 75.00 \\
    \midrule
    \rowcolor{gray!15}
    \multicolumn{3}{c}{\textit{Visual-Only}} \\
    \midrule
    OmniVinci & 65.62\rlap{\textsubscript{\color{green}-12.50}} & 40.62\rlap{\textsubscript{\color{green}-28.13}} \\
    MiniCPM-o 4.5 & 78.12\rlap{\textsubscript{\color{green}-6.26}} & 46.88\rlap{\textsubscript{\color{green}-28.12}} \\
    \midrule
    \rowcolor{gray!15}
    \multicolumn{3}{c}{\textit{Audio-Only}} \\
    \midrule
    OmniVinci & 75.00\rlap{\textsubscript{\color{green}-3.12}} & 62.50\rlap{\textsubscript{\color{green}-6.25}} \\
    MiniCPM-o 4.5 & 84.38\rlap{\textsubscript{\color{green}-0.00}} & 68.75\rlap{\textsubscript{\color{green}-6.25}} \\
    \bottomrule
  \end{tabular}
  }
\end{wraptable}
\textbf{Contextual Breadth.} The script-based approach inherently captures broader context. We compare the average temporal span of both sets of QA pairs, calculated by subtracting the end timestamp of the first relevant segment from the start timestamp of the last.
As shown in Table~\ref{tab:time}, the average temporal span for the script-based approach is 37.88s higher than that of the video-based baseline. The results across different durations show that the advantage of integrating long-term information is more significant in longer videos, with the gap reaching 79.65s for videos ranging from 8 to 14 minutes.
% This makes the script-based QA pairs more challenging for models. 

\textbf{Cross-Modal Synergy.} This increased difficulty of script-based QA is corroborated by the results in Table~\ref{tab:modal}, where we evaluate the performance of MLLMs specifically on the 8--14min samples. Across all modality settings, models exhibit a consistent performance drop when switching from video-based to script-based QA pairs. Moreover, the performance degradation observed when removing a single modality is more severe for script-based QA. This disparity shows that our pipeline can generate more rigorous QA pairs that necessitate cross-modal synergy.

% \textbf{Script vs Omni-modal Caption}
% 类似OmniVinci，视频分割成20s每段，分别生成音频和视觉描述，利用LLM为每段生成Omni-Caption。
% 各生成73条成对比较选出更好的qa对。基于Script的胜47次，Caption胜26次

\subsubsection{Impact of Data Scaling}
\begin{table}[!htbp]
\centering
\caption{Performance scaling of Qwen2.5-Omni-7B across different data volumes. ``Avg.'' denotes the average score across all listed benchmarks.}
\label{tab:data}
\resizebox{\linewidth}{!}{
\begin{tabular}{l|c|cccccccc}
\toprule
\textbf{Data} & \textbf{Avg.} & \textbf{Video-MME}\textsubscript{short} & \textbf{Video-MME-v2} & \textbf{Daily-Omni} & \textbf{OmniVideoBench} & \textbf{JointAVBench} & \textbf{FutureOmni} & \textbf{OmniVideo-Test} \\
\midrule
w/o SFT & 47.26 & 75.56 & 10.28 & 62.41 & 36.54 & 54.44 & 48.85 & 42.77 \\
OmniVideo-10K & 52.64 & \textbf{77.22} & \textbf{13.84} & 69.92 & 39.10 & 60.57 & 52.60 & 55.25 \\
OmniVideo-25K & 53.98 & 76.56 & 13.74 & 70.68 & 42.63 & 60.15 & 54.27 & 59.80 \\
OmniVideo-50K & 54.14 & 76.33 & 11.38 & 69.84 & \textbf{42.83} & 59.45 & \textbf{57.60} & 61.58 \\
OmniVideo-75K & \textbf{54.32} & 76.89 & 10.47 & \textbf{72.26} & 41.26 & \textbf{61.03} & 55.73 & \textbf{62.57} \\
OmniVideo-100K & 52.98 & 76.33 & 8.50 & 69.84 & 39.88 & 60.75 & 55.00 & 60.59 \\
\bottomrule
\end{tabular}
}
\end{table}

To investigate how the volume of our generated data affects model performance, we fine-tune Qwen2.5-Omni on different subsets of OmniVideo ranging from 10K to 100K samples. All training settings remain consistent with those used in the main experiments. As shown in Table~\ref{tab:data}, compared to the baseline model (w/o SFT), incorporating merely 10K samples leads to a performance leap across almost all benchmarks. For instance, accuracy on Daily-Omni and JointAVBench jumps from 62.41\% to 69.92\% and 54.44\% to 60.57\%, respectively. Performance exhibits steady improvements as the data volume increases from 10K to 75K, with the average score (Avg.) peaking at 54.32\% at the 75K mark. However, when the data volume is further expanded to 100K, we observe slight performance fluctuations or saturation across several metrics. Nonetheless, the overall scaling trajectory validates the effectiveness and efficiency of our automated data generation pipeline.

\section{Related Work}
\textbf{Multimodal Large Language Models.}
Building on the success of Multimodal Large Language Models~\citep{llava-onevision,qwen2.5-vl,smolvlm,internvl3.5,cap4video++,selongvlm,momentor++}, researchers~\citep{anygpt,videollama2,vita,mini-omni2,vita-1.5,baichuan-omni-1.5,ming-omni} expand model capabilities from visual-only analysis to audio-visual understanding by adding audio branches. These models typically concatenate visual and audio tokens directly before feeding them into the LLM. To better capture the synchronization within audio-visual videos, several works~\citep{qwen2.5-omni,video-salmonn2,omnivinci,uni-moe-2.0-omni} introduce specialized designs for temporal alignment. For example, Qwen3-Omni~\citep{qwen3-omni} interleaves visual and audio tokens based on actual time. JavisGPT~\citep{javisinst-und_javisgpt} employs a SyncFusion module to integrate audio information into visual features, explicitly modeling spatio-temporal synchrony.

\textbf{Audio-Visual Video Datasets.}
Early studies~\citep{avsd,pano-avqa,music-avqa,avqa,music-avqa-v2.0} establish the foundation for audio-visual question answering, followed by a surge of datasets and benchmarks~\citep{valor-32k,worldsense,omnivideobench,socialomni,lvomnibench,mmou,avinstruct,internvid2,avhbench,omnieval,avut,javisinst-und_javisgpt,futureomni,egoavu,trisense-2m} designed to enhance and evaluate the cross-modal capabilities of MLLMs across diverse scenarios. Most recent works~\citep{longvale,jointavbench,omnivinci,longshotbench} employing automated data engines typically segment videos and use specialized models to extract uni-modal information (such as visual captions, speech transcriptions, and audio descriptions). These texts are then jointly fed into LLMs to produce segment-level audio-visual captions, which are ultimately used to generate QA pairs. However, this method often leads to narrative incoherence. Although Daily-Omni~\citep{daily-omni} attempts to address this by using the full video to revise and align descriptions across modalities, this post-process may fail to preserve all original details, leading to information loss.

\textbf{Video Structured Scripting.} Conventional video captioning typically provides a single descriptive text for each segment. Some efforts~\citep{vc4vg, asid} have explored multi-attribute or multi-dimensional structuring to yield more fine-grained captions. More recently, TimeChat-Captioner~\citep{timechatcaptioner} proposes the Omni Dense Captioning task, generating script-like captions by utilizing a six-dimensional schema to ensure comprehensive coverage of audio-visual content. However, descriptions generated by such methods often remain logically isolated across segments. In contrast, we introduce a main entity list as a global prior to synthesize coherent audio-visual scripts. Among concurrent works, OmniScript~\citep{omniscript} focuses on training MLLMs to generate scripts directly from video, while Script-a-Video~\citep{scriptavideo} adopts the script as a bridge that distills video content into structured evidence for understanding and provides a blueprint for video generation. Distinct from these works, we leverage structured scripts to generate audio-visual QA pairs with long-term temporal spans and cross-modal dependencies. These samples are used to fine-tune MLLMs, thereby enhancing their comprehension capabilities in audio-visual scenarios.

\section{Conclusion}
In this paper, we propose an automated data synthesis engine designed to enhance the audio-visual understanding capabilities of MLLMs. Our pipeline is built upon two stages: (1) We extract audio and visual information from video segments to construct a structured, script-like text, leveraging the main entity list prior to maintain narrative coherence across segments and explicitly reconstruct audio-visual associations. (2) We design a clue-guided QA generation strategy that prompts the model to first mine clues involving long-term temporal spans and cross-modal dependencies, and then generate QA pairs based on these clues. The resulting instruction-tuning dataset, OmniVideo-100K, effectively addresses the shortcomings of existing open-source models in synergistic cross-modal comprehension, demonstrating consistent performance gains on both our test set, OmniVideo-Test, and established benchmarks.
In future work, we plan to explore more robust adaptive segmentation strategies and integrate specialized audio models to capture fine-grained sounds and paralinguistic features, thereby further elevating the quality of automatically generated data.

\subsubsection*{Acknowledgments}
This work is funded by Fundamental and Interdisciplinary Disciplines Breakthrough Plan of the Ministry of Education of China
(JYB2025XDXM902).

\bibliographystyle{iclr2026_conference}
\bibliography{iclr2026_conference}

\newpage
\appendix
\section{Audio-Visual Task Taxonomy}
\label{sec:tasks}
\begin{table}[!htbp]
\centering
\caption{Ten audio-visual tasks covered by OmniVideo-100K and OmniVideo-Test.}
\label{tab:chinese_taxonomy}
\resizebox{\linewidth}{!}{
\begin{tabular}{l|l|l}
\toprule
\textbf{Cognitive Level} & \textbf{Task} & \textbf{Definition} \\
\midrule
\multirow{2}{*}{\textbf{Alignment}} & Fine-Grained Perception & Perceive synchronized cross-modal details from given unimodal cues. \\
\cmidrule{2-3}
 & Scene Transformation Detection & Identify synchronized visual shifts using audio cues. \\
\midrule
\multirow{5}{*}{\textbf{Understanding}} & Context Understanding & Identify semantic associations between audio and visual elements. \\
\cmidrule{2-3}
 & Comparison & Compare target states or attributes across different timestamps. \\
\cmidrule{2-3}
 & Sentiment Analysis & Analyze character inner states via audio-visual cues. \\
\cmidrule{2-3}
 & Event Sequence Ordering & Restore the correct temporal order of jumbled events. \\
\cmidrule{2-3}
 & Summarization & Generate an audio-visual summary for specific events. \\
\midrule
\multirow{3}{*}{\textbf{Reasoning}} & Causal Reasoning & Infer underlying causes of specific events. \\
\cmidrule{2-3}
 & Future Prediction & Predict upcoming plot developments from current content. \\
\cmidrule{2-3}
 & Hypothetical Reasoning & Propose ``What-if'' scenarios to re-interpret video content. \\
\bottomrule
\end{tabular}
}
\end{table}

\section{Detailed Experimental Settings and Results}
\begin{table}[!htbp]
\centering
\caption{Parameter settings for full-parameter fine-tuning of VITA-1.5, Qwen2.5-Omni-7B and Qwen3-Omni-30B-A3B-Instruct.}
\label{tab:args}
\resizebox{\linewidth}{!}{
\begin{tabular}{l|ccccccc}
\toprule
\textbf{Models} & \textbf{Max Pixels} & \textbf{FPS} & \textbf{Max Frames} & \textbf{Epochs} & \textbf{Batch Size} & \textbf{Learning Rate} & \textbf{Warmup Ratio}\\
\midrule
VITA-1.5 & default & default & default & 1 & 32 & $1\times10^{-5}$ & 0.03 \\
Qwen2.5-Omni & $448\times448$ & 1.0 & 180 & 1 & 32 & $1\times10^{-5}$ & 0.1\\
Qwen3-Omni & $256\times256$ & 1.0 & 150 & 1 & 32 & $5\times10^{-6}$ & 0.1\\
\bottomrule
\end{tabular}
}
\end{table}

\begin{table}[!htbp]
\centering
\caption{Performance comparison on OmniVideo-Test. FGP: Fine Grained Perception, STD: Scene Transformation Detection, CU: Context Understanding, CP: Comparison, SA: Sentiment Analysis, ESO: Event Sequence Ordering, SM: Summarization, CR: Causal Reasoning, FP: Future Prediction, HR: Hypothetical Reasoning.}
\label{tab:avs-bench_results_details}
\resizebox{\linewidth}{!}{
\begin{tabular}{lc|c|cccccccccc}
\toprule
\textbf{Models} & \textbf{Size} & \textbf{Overall} & \textbf{FGP} & \textbf{STD} & \textbf{CU} & \textbf{CP} & \textbf{SA} & \textbf{ESO} & \textbf{SM} & \textbf{CR} & \textbf{FP} & \textbf{HR}\\
\midrule
Gemini-3.1-Pro & - & 83.96 & 78.33 & 89.29 & 87.04 & 83.64 & 74.55 & 86.67 & 94.12 & 82.69 & 70.27 & 95.24 \\
\midrule
% \multicolumn{13}{c}{\textit{Open Source Models}} \\ 
% \midrule
VITA-1.5 & 7B & 40.99 & 30.00 & 26.79 & 46.30 & 49.09 & 47.27 & 35.00 & 35.29 & 48.08 & 43.24 & 52.38 \\
Qwen2.5-Omni & 7B & 42.77 & 38.33 & 41.07 & 40.74 & 58.18 & 40.00 & 45.00 & 50.00 & 34.62 & 35.14 & 45.24 \\
video-SALMONN 2+ & 7B & 45.15 & 45.00 & 30.36 & 51.85 & 50.91 & 41.82 & 46.67 & 58.82 & 34.62 & 45.95 & 52.38 \\
OmniVinci & 9B & 47.13 & 45.00 & 42.86 & 53.70 & 56.36 & 49.09 & 43.33 & 58.82 & 38.46 & 32.43 & 52.38 \\
MiniCPM-o 4.5 & 9B & 55.25 & 55.00 & 58.93 & 74.07 & 50.91 & 43.64 & 61.67 & 61.76 & 40.38 & 54.05 & 52.38 \\
Qwen3-Omni & 30B & 49.70 & 40.00 & 46.43 & 59.26 & 65.45 & 52.73 & 45.00 & 52.49 & 48.08 & 37.84 & 47.62 \\
Uni-MoE-2.0-Omni & 30B & 46.93 & 48.33 & 30.36 & 68.52 & 69.09 & 45.45 & 30.00 & 52.94 & 48.08 & 43.24 & 33.33 \\
\midrule
% \multicolumn{13}{c}{\textit{MLLMs trained with OmniVideo-100K}} \\ 
% \midrule
OmniVideo-7B\textsubscript{(VITA-1.5)} & 7B & 61.58 & 55.00 & 64.29 & 64.81 & 65.45 & 63.64 & 53.33 & 73.53 & 65.38 & 62.16 & 52.38 \\
OmniVideo-7B\textsubscript{(Qwen2.5-Omni)} & 7B & 60.59 & 61.67 & 64.29 & 61.11 & 63.64 & 63.64 & 60.00 & 64.71 & 50.00 & 56.76 & 59.52 \\
OmniVideo-30B\textsubscript{(Qwen3-Omni)} & 30B & 63.56 & 60.00 & 60.71 & 70.37 & 67.27 & 65.45 & 61.67 & 73.53 & 57.69 & 59.46 & 61.90 \\
% \midrule
% \multicolumn{13}{c}{\textit{Audio-Only}} \\
% \midrule
% MiniCPM-o 4.5 & 9B & 45.74 & 38.33 & 33.93 & 55.56 & 56.36 & 40.00 & 51.67 & 58.82 & 42.31 & 43.24 & 40.48 \\
% Qwen3-Omni & 30B & 46.14 & 43.33 & 30.36 & 50.00 & 58.18 & 54.55 & 45.00 & 58.82 & 46.15 & 32.43 & 42.86 \\
% \midrule
% \multicolumn{13}{c}{\textit{Visual-Only}} \\
% \midrule
% MiniCPM-o 4.5 & 9B & 47.92 & 45.00 & 39.29 & 53.70 & 50.91 & 41.82 & 51.67 & 52.94 & 51.92 & 43.24 & 50.00 \\
% Qwen3-Omni & 30B & 47.13 & 46.67 & 37.50 & 50.00 & 54.55 & 43.64 & 43.33 & 55.88 & 51.92 & 40.54 & 50.00 \\
\bottomrule
\end{tabular}
}
\end{table}

\newpage
\section{Prompts for the Overall Pipeline}
\label{sec:prompts}

\begin{figure}[!htbp]
    \centering
    \includegraphics[trim=0 490 275 0, clip, width=1.0\textwidth]{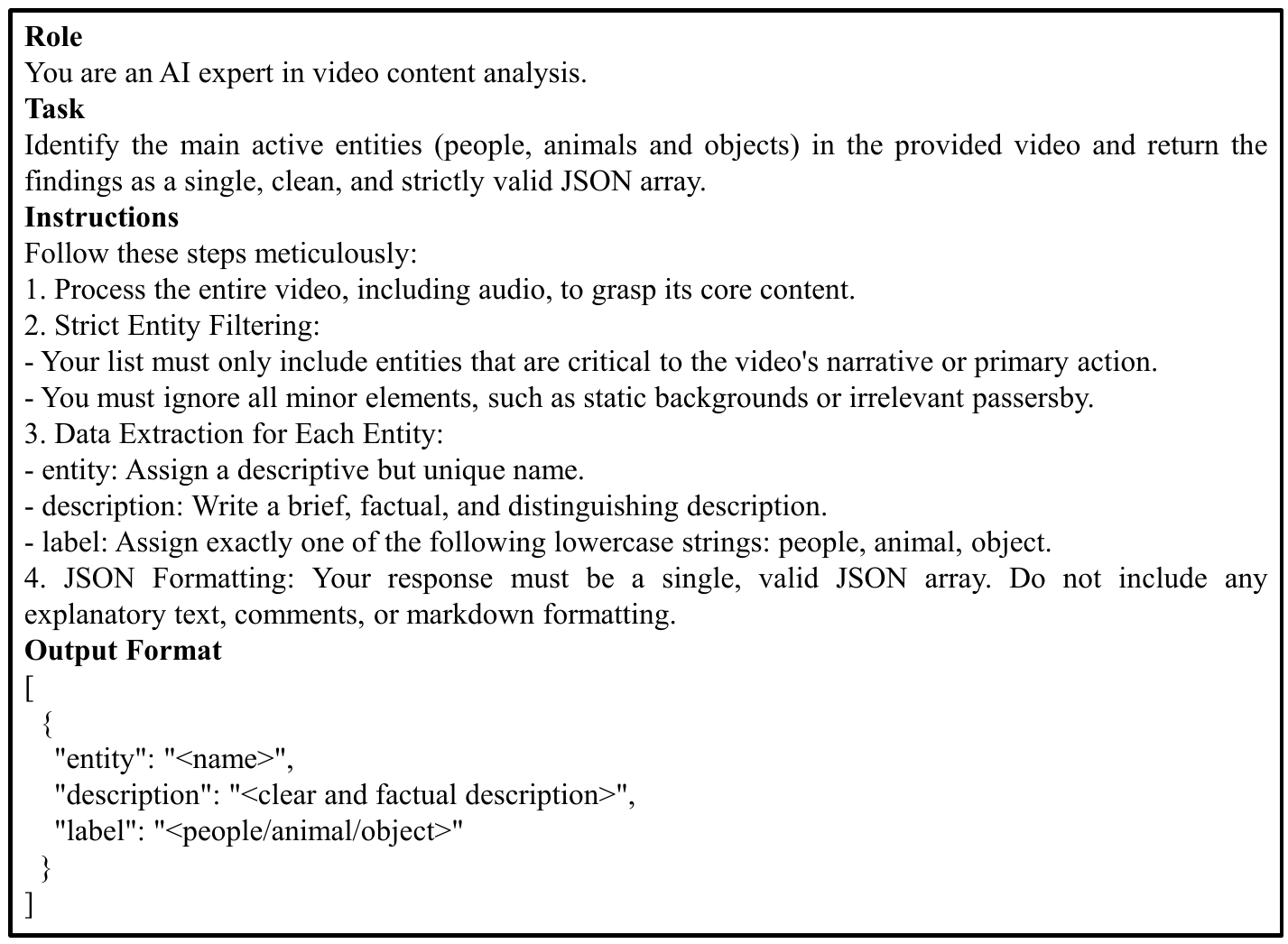}
    \caption{Prompt for instructing MLLMs to identify main active entities in the video.}
\end{figure}

\begin{figure}[!htbp]
    \centering
    \includegraphics[trim=0 745 275 0, clip, width=1.0\textwidth]{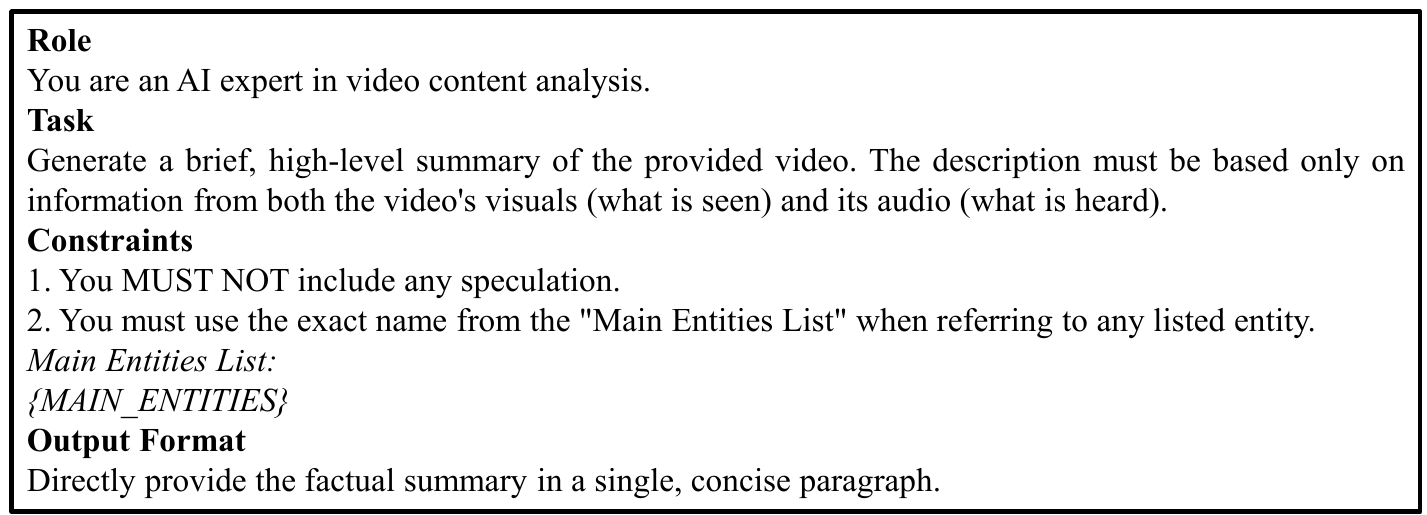}
    \caption{Prompt for instructing MLLMs to generate a video summary.}
\end{figure}

\begin{figure}[!htbp]
    \centering
    \includegraphics[trim=0 470 275 0, clip, width=1.0\textwidth]{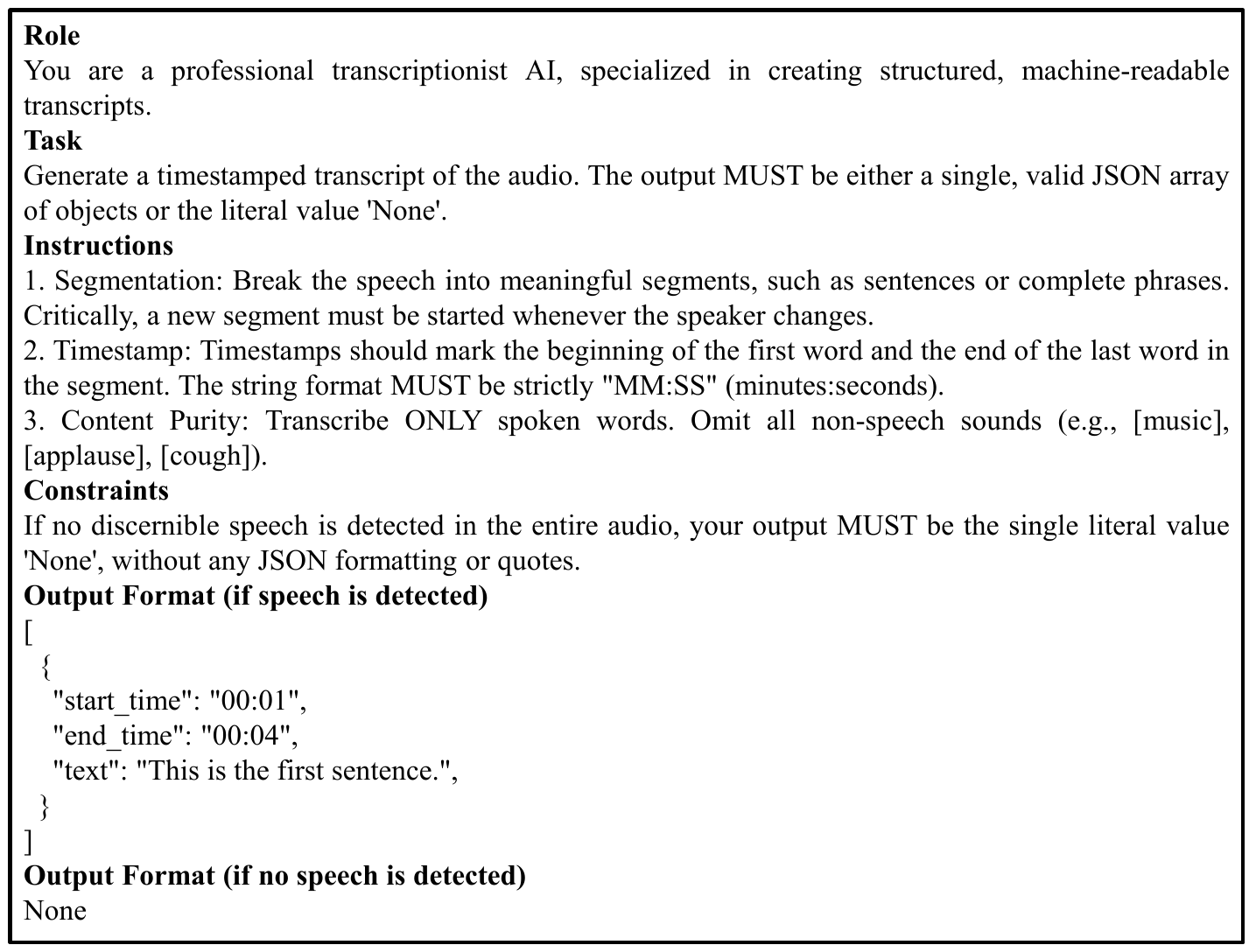}
    \caption{Prompt for instructing MLLMs to generate transcriptions for speech.}
\end{figure}

\begin{figure}[!htbp]
    \centering
    \includegraphics[trim=0 255 275 0, clip, width=1.0\textwidth]{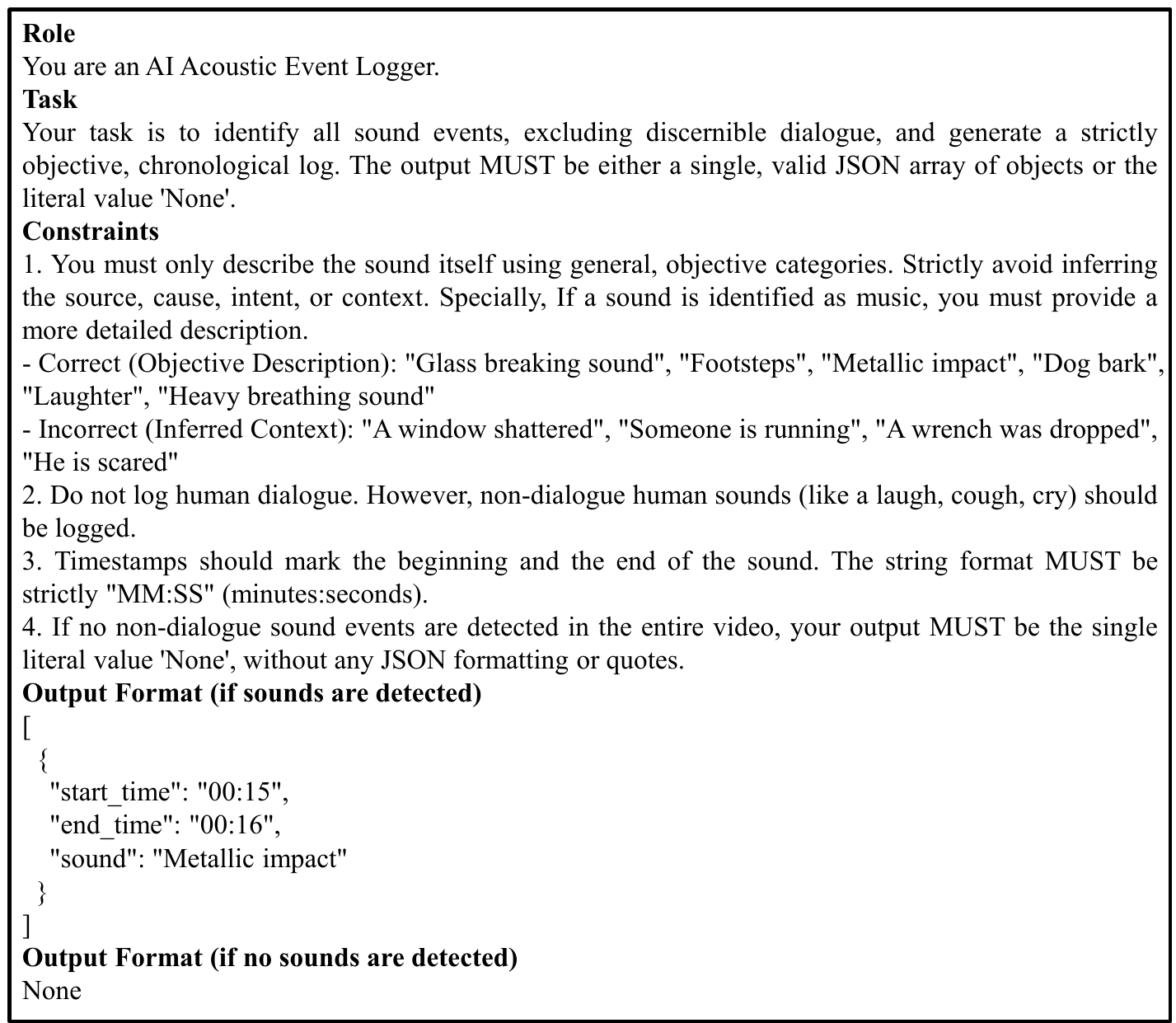}
    \caption{Prompt for instructing MLLMs to identify non-speech sounds.}
\end{figure}

\begin{figure}[!htbp]
    \centering
    \includegraphics[trim=0 0 275 0, clip, width=1.0\textwidth]{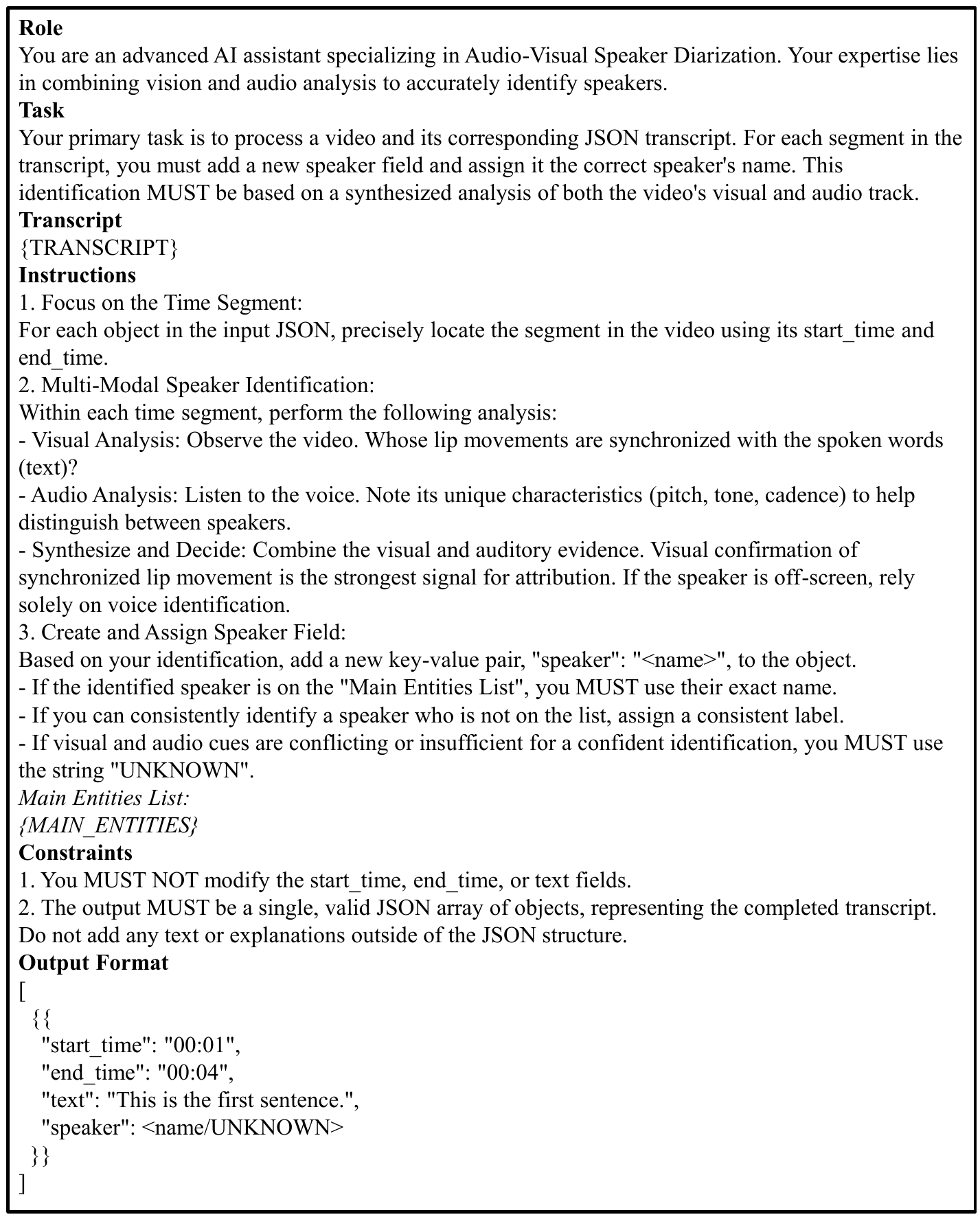}
    \caption{Prompt for instructing MLLMs to identify speakers based on audio-visual cues.}
\end{figure}

\begin{figure}[!htbp]
    \centering
    \includegraphics[trim=0 293 275 0, clip, width=1.0\textwidth]{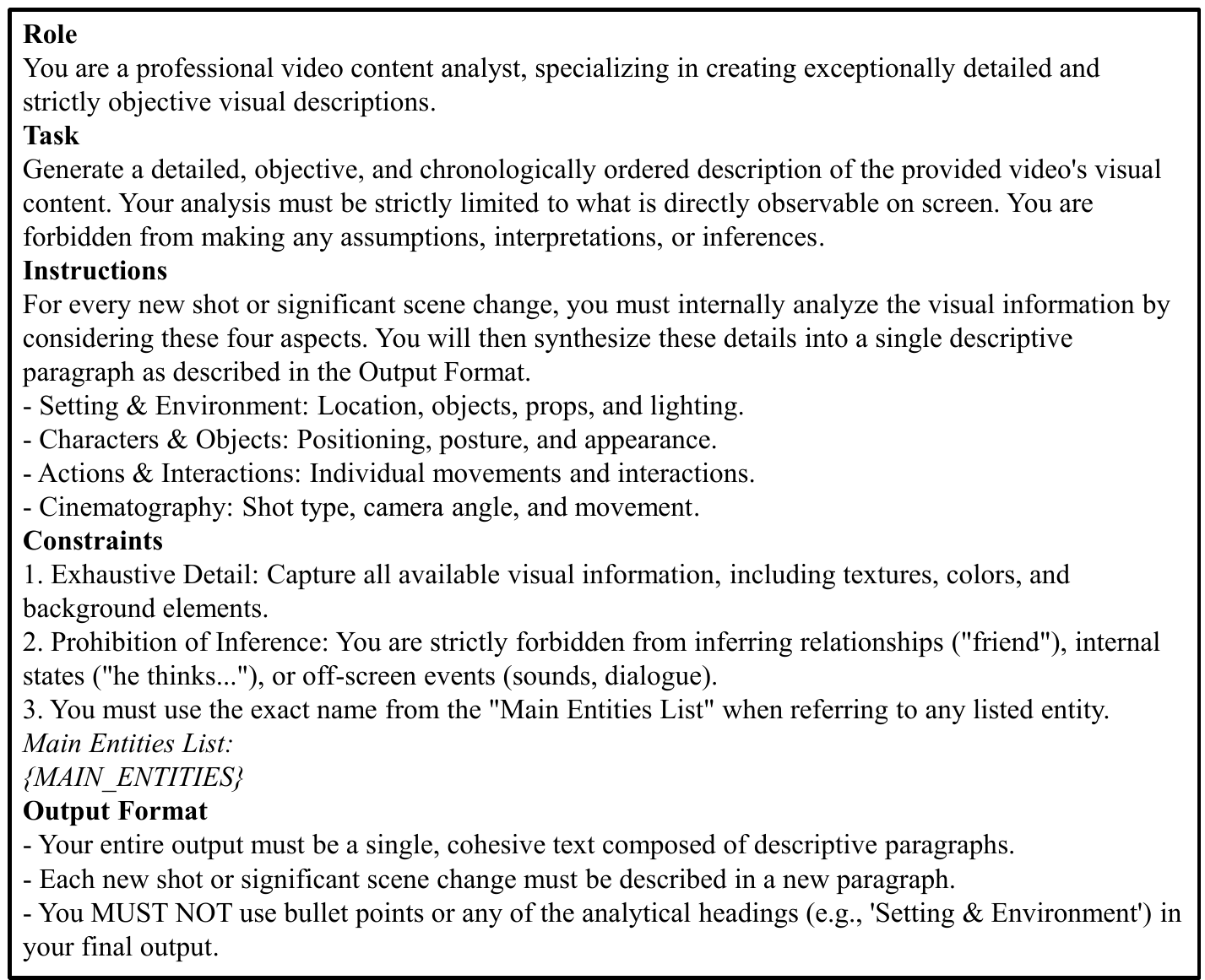}
    \caption{Prompt for instructing MLLMs to generate visual descriptions.}
\end{figure}

\begin{figure}[!htbp]
    \centering
    \includegraphics[trim=0 45 275 0, clip, width=1.0\textwidth]{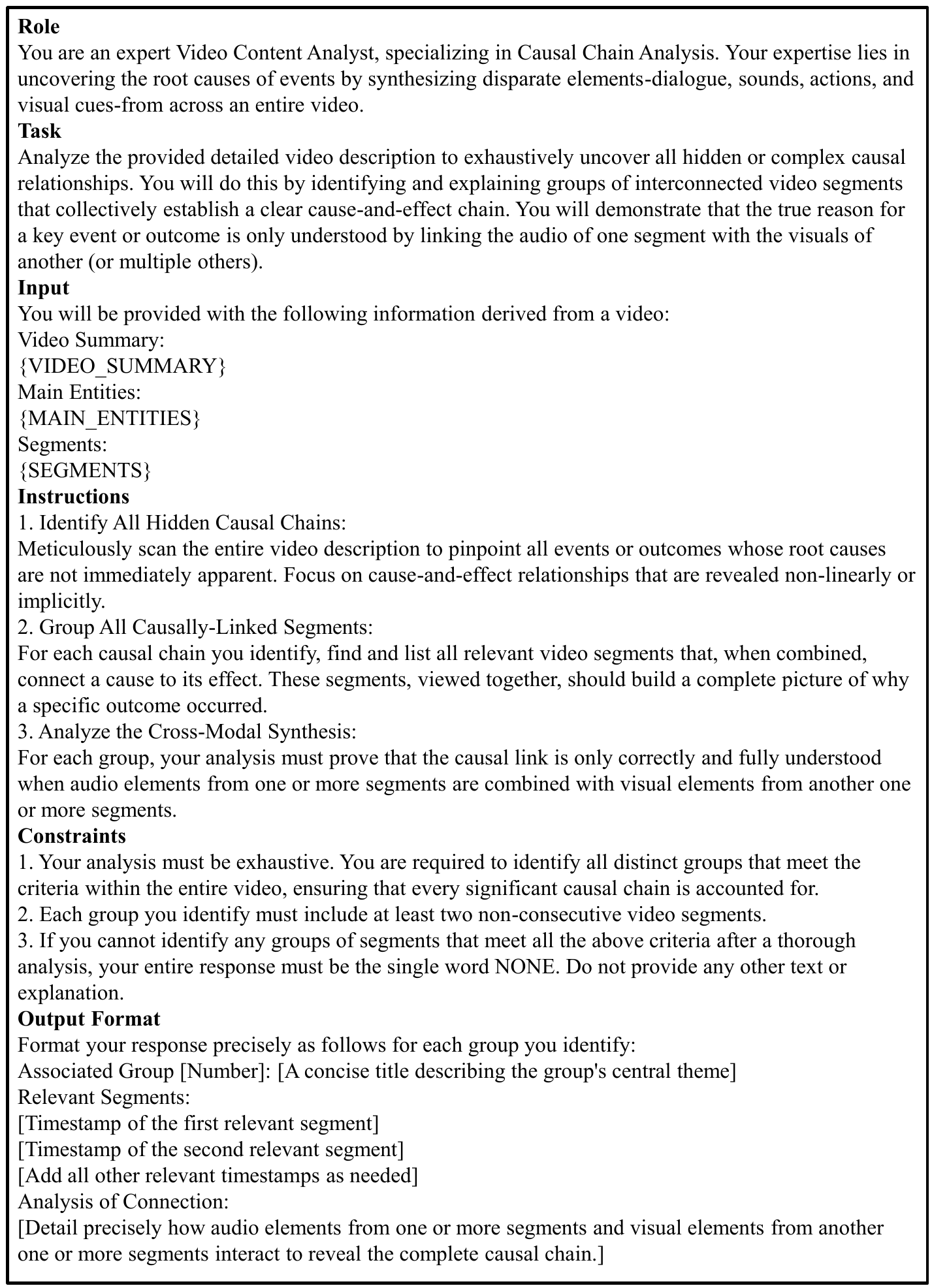}
    \caption{Prompt for Global Clue Mining of Clue-Guided Generation (Causal Reasoning).}
\end{figure}

\begin{figure}[!htbp]
    \centering
    \includegraphics[trim=0 140 275 0, clip, width=1.0\textwidth]{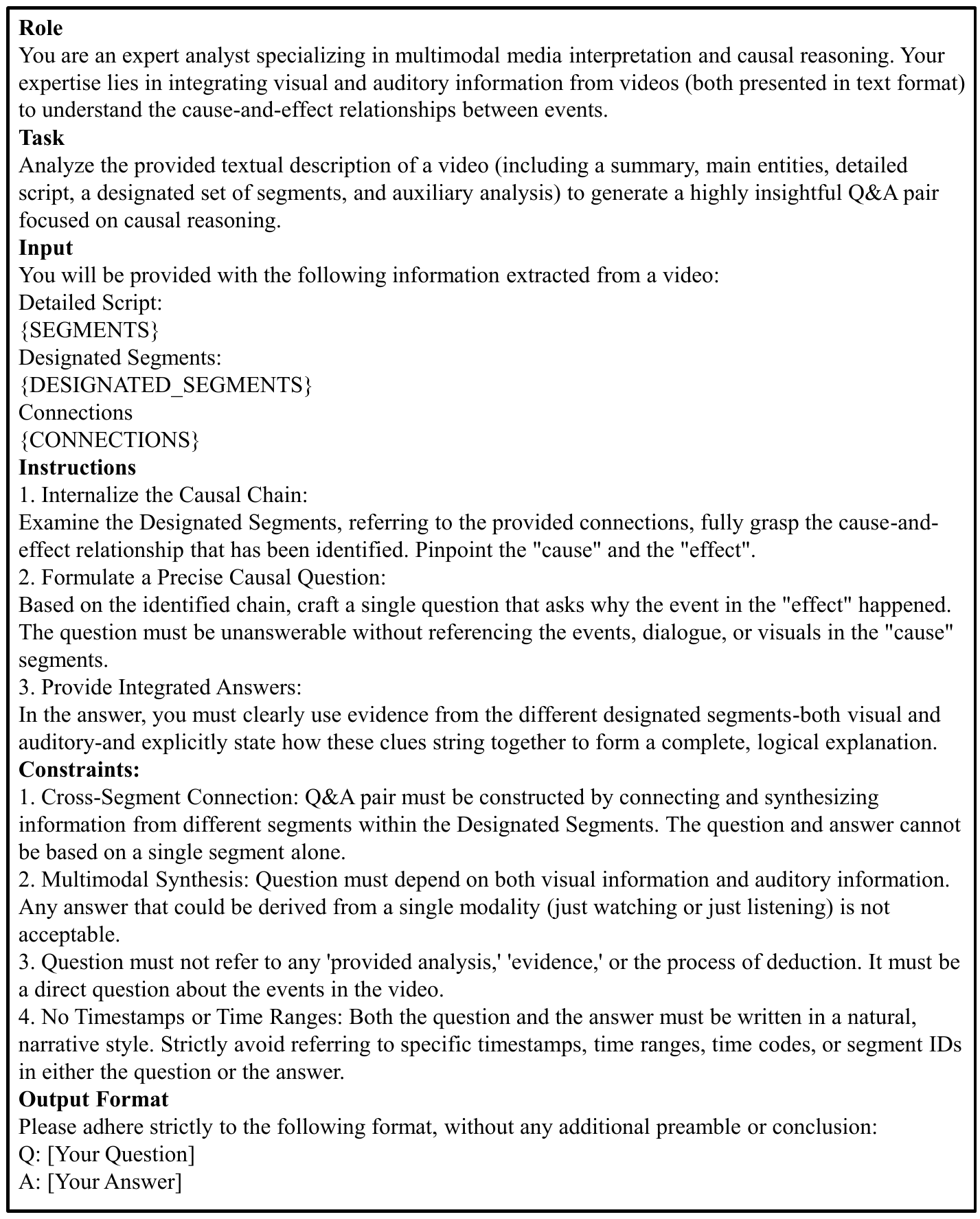}
    \caption{Prompt for Locally Focused Generation of Clue-Guided Generation: generating open-ended QA pairs (Causal Reasoning).}
\end{figure}

\begin{figure}[!htbp]
    \centering
    \includegraphics[trim=0 193 275 0, clip, width=1.0\textwidth]{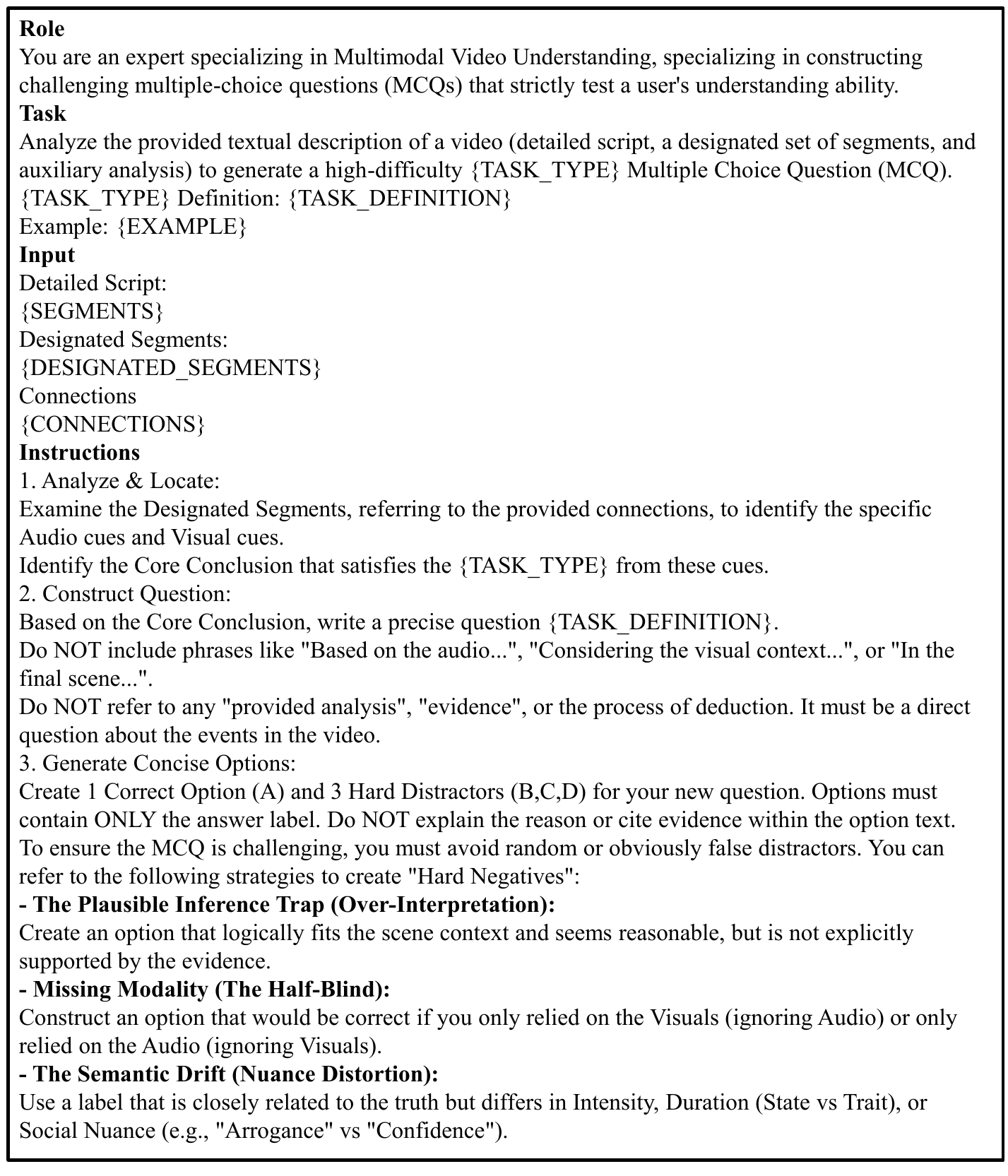}
    \caption{Prompt for Locally Focused Generation of Clue-Guided Generation: generating multiple-choice questions.}
\end{figure}

\begin{figure}[!htbp]
    \ContinuedFloat
    \centering
    \includegraphics[trim=0 318 275 0, clip, width=1.0\textwidth]{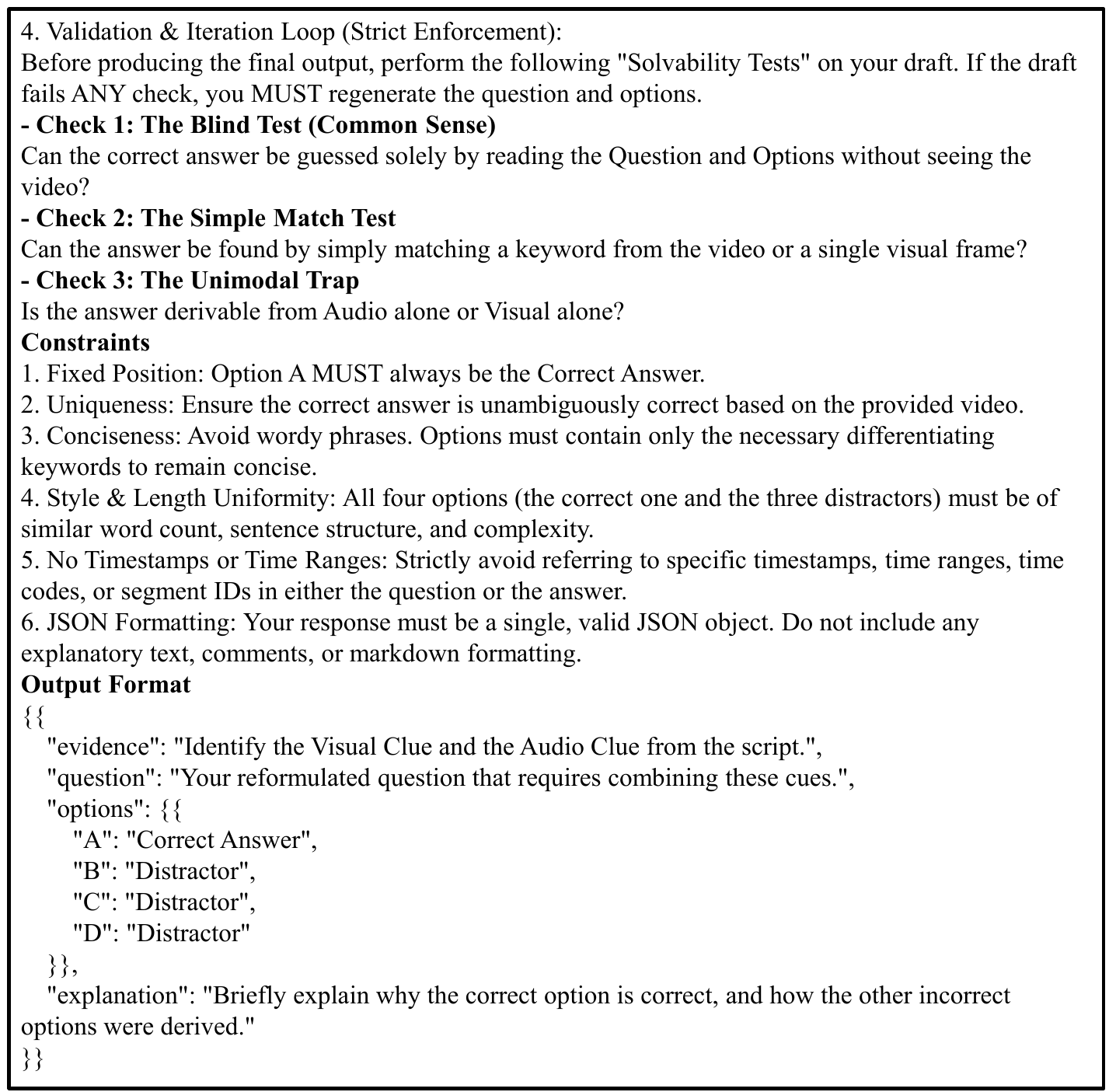}
    \caption{Prompt for Locally Focused Generation of Clue-Guided Generation: generating multiple-choice questions (continued).}
\end{figure}

\end{document}